%% file: main.tex
\documentclass[10pt,twocolumn,letterpaper]{article}

\usepackage[pagenumbers]{cvpr} 

\definecolor{cvprblue}{rgb}{0.21,0.49,0.74}
\usepackage[pagebackref,breaklinks,colorlinks,citecolor=cvprblue]{hyperref}
\usepackage{bm}
\usepackage[accsupp]{axessibility}

\usepackage{multirow}
\usepackage{booktabs}
\usepackage{array}
\usepackage{graphicx}  
\usepackage{gensymb}
\usepackage{physics}
\usepackage{makecell}
\usepackage{xcolor}  

\def\paperID{*****} 
\def\confName{CVPR}
\def\confYear{2025}

\title{Relative Pose Estimation through Affine Corrections of Monocular Depth Priors}

\author{Yifan Yu$^{1,2}$\thanks{This work was primarily done as Yifan's master thesis at ETH Zurich.} \quad
Shaohui Liu$^{1}$ \quad
R\'emi Pautrat$^{3}$ \quad
Marc Pollefeys$^{1,3}$ \quad
Viktor Larsson$^{4}$ \\
$^{1}$ETH Zurich \quad $^{2}$PICO \quad $^{3}$Microsoft Spatial AI Lab \quad $^{4}$Lund University \\
}

\begin{document}
\maketitle
\input{sec/0_abstract_v1}    
\input{sec/1_intro_v1}
\input{sec/2_related_work}
\input{sec/3_problem_formulation}
\input{sec/4_method}
\input{sec/5_exp}

\input{sec/6_conclusion}

{
    \small
    \bibliographystyle{ieeenat_fullname}
    \bibliography{reference}
}
\clearpage
\input{sec/X_suppl}

\end{document}

%% file: sec/0_abstract_v1.tex
\begin{abstract}
Monocular depth estimation (MDE) models have undergone significant advancements over recent years. Many MDE models aim to predict affine-invariant relative depth from monocular images, while recent developments in large-scale training and vision foundation models enable reasonable estimation of metric (absolute) depth.
However, effectively leveraging these predictions for geometric vision tasks, in particular relative pose estimation, remains relatively under explored. While depths provide rich constraints for cross-view image alignment, the intrinsic noise and ambiguity from the monocular depth priors present practical challenges to improving upon classic keypoint-based solutions.
In this paper, we develop three solvers for relative pose estimation that explicitly account for independent affine (scale and shift) ambiguities, covering both calibrated and uncalibrated conditions. 
We further propose a hybrid estimation pipeline that combines our proposed solvers with classic point-based solvers and epipolar constraints. 
We find that the affine correction modeling is beneficial to not only the relative depth priors but also, surprisingly, the ``metric" ones. Results across multiple datasets demonstrate large improvements of our approach over classic keypoint-based baselines and PnP-based solutions, under both calibrated and uncalibrated setups. We also show that our method improves consistently with different feature matchers and MDE models, and can further benefit from very recent advances on both modules. Code is available at \url{https://github.com/MarkYu98/madpose}.
\end{abstract}
\vspace{-9pt}

%% file: sec/1_intro_v1.tex
\section{Introduction}
\label{sec:intro}

Recently, deep learning based monocular depth estimation (MDE) has made remarkable progress~\cite{birkl2023midas3.1, kar20223d-omnidata, marigold, depthanything, depth_anything_v2, wang2024moge}, offering increasingly accurate 3D information from single 2D images. These advancements have opened up new possibilities for enriching traditional geometric computer vision tasks with 3D priors. The ability to infer depth information from a single image has been shown to have useful implications in various applications \cite{czarnowski2020deepfactors,packnet,shan2023animating}.

Despite the significant progress in monocular depth estimation, the integration of these depth priors in fundamental geometric computer vision tasks, particularly camera pose estimation, remains under-studied. While incorporating depth information would be beneficial intuitively, effectively leveraging these priors to find geometric relationships between multiple views presents unique challenges that have not been fully addressed by existing works~\cite{studiosfm, barath2022relative, ding2024fundamental}.

\input{figures/formulation}

A critical limitation of existing approaches is the common assumption that the predicted depth maps from different views can be related by a single scale factor. This assumption, however, fails to account for the intrinsic properties of existing monocular depth estimation models, as state-of-the-art MDE models~\cite{marigold, depth_anything_v2, wang2024moge} are typically trained to predict relative depth or disparity (inverse-depth) that are invariant to both scale and shift (affine) transformations. While there have been recent progress in developing models for metric depth estimation \cite{bhat2023zoedepth,piccinelli2024unidepth,yin2023metric3d,depth_anything_v2}, we find that surprisingly, they also benefit from modeling the affine corrections as they are not perfectly consistent with the actual metric depth (See Figure \ref{fig:gt-shift}, Table \ref{tab:results-calib-scannet}, and \cref{sec:model_scale_and_shift}). 

In this paper, we propose three solvers specialized for solving relative pose under calibrated or uncalibrated camera setups. These solvers use pixel correspondences along with their depth priors from MDE models as input, and incorporate explicit modeling of both scale and shift variations in the depth predictions. Specifically, the solvers introduced are as follows (with only the calibrated solver being minimal and the two others being over-constrained):
\begin{itemize} \itemsep0pt
    \item \textbf{Calibrated 3-point solver}: for calibrated image pairs. 
    \item \textbf{Shared-focal 4-point solver}: for uncalibrated image pairs with a common unknown focal length.
    \item \textbf{Two-focal  4-point solver}: for uncalibrated image pairs.
\end{itemize}

In addition, we integrate these new solvers into a flexible hybrid robust estimation pipeline that combines our depth-aware solvers with classic point-based solvers~\cite{Nister03FivePt,hartley_zisserman_2004,Stewenius2008}. We also develop hybrid solutions for scoring and local optimization, where we optimize both the classic Sampson error~\cite{Sampson1982FittingCS} and depth-induced reprojection error using depth and the solved affine corrections. This enables combining the advantages of both worlds and results in a robust relative pose estimator that achieves consistent improvement across datasets. Our key contributions are summarized as follows:
\begin{itemize}
    \item We propose to solve relative pose with explicit affine (scale and shift) corrections on monocular depth predictions, addressing a limitation in existing methods.
    \item We develop 3 new solvers tailored to different calibration setups: calibrated, uncalibrated with common focal length, and fully uncalibrated image pairs.
    \item Our hybrid estimation pipeline integrates depth-aware and classic point-based solvers, scoring and local optimization, leading to largely improved accuracy and robustness in relative pose estimation.
    \item Our framework is compatible with a wide range of image matchers and MDE models with consistent improvement, making it easy to integrate in existing pipelines.
\end{itemize}

%% file: figures/formulation.tex
\begin{figure}[tb]
    \centering
    \includegraphics[width=0.95\linewidth]{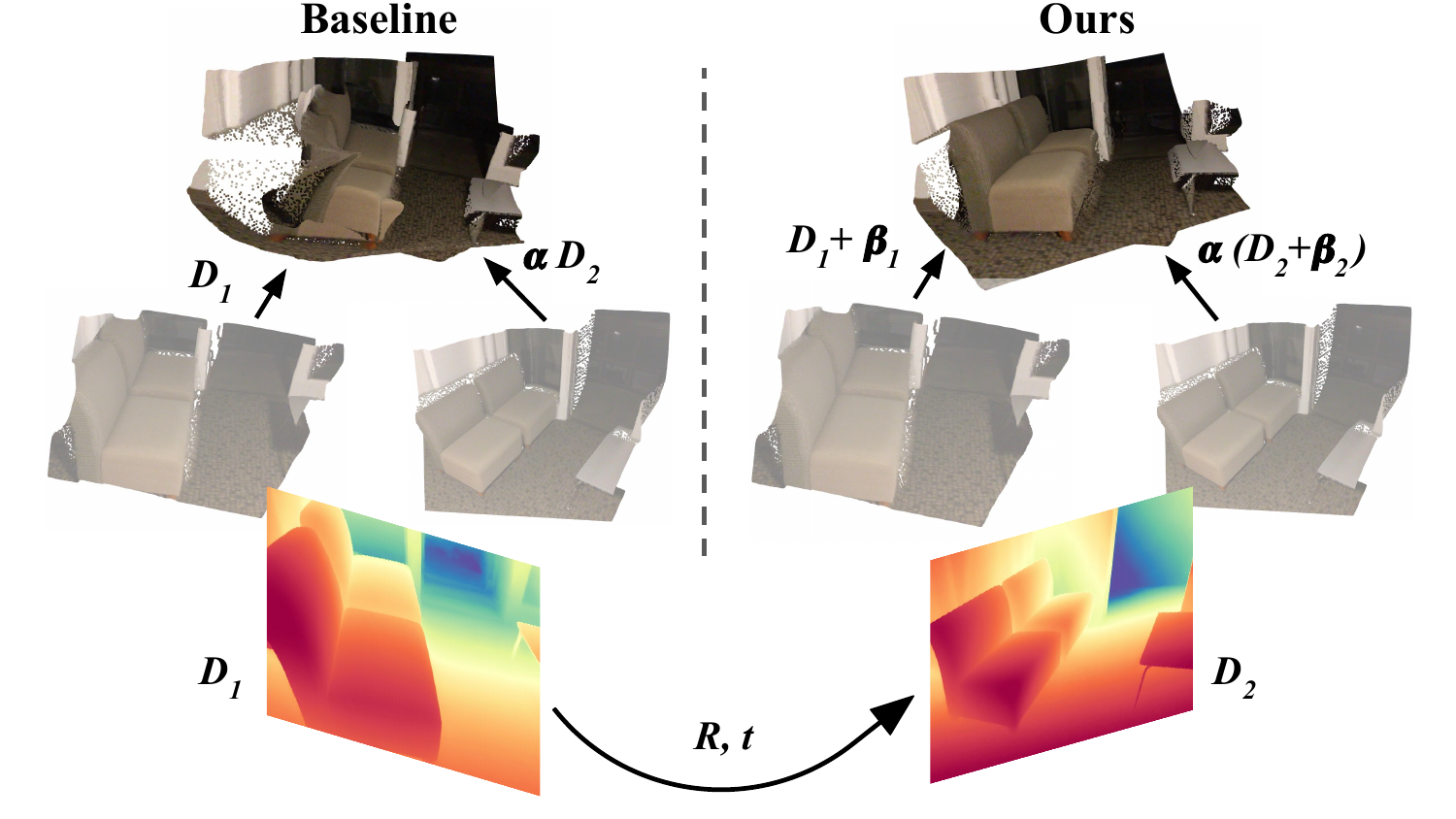}
    \caption{Our method jointly estimates affine corrections of monocular depth maps $D_1 + \beta_1$ and $\alpha (D_2 + \beta_2)$ together with relative pose $\bm{R}, \bm{t}$ (\textbf{Right}), whereas the classic way of aligning the depth maps with only scale modeling ($\alpha$) leads to wrong and distorted alignments (\textbf{Left}).}
    \label{fig:formulation}
\end{figure}

%% file: sec/2_related_work.tex
\section{Related Work}
\label{sec:related_work}

\noindent\textbf{Monocular Depth Estimation:}
Recovering depth information from a single image is a fundamental problem in 3D computer vision. Early methods \cite{saxena2005learning,saxena2008make3d} resort to probabilistic modeling, and the advent of deep learning enables a variety of robust monocular depth estimators \cite{eigen2014depthsingleimage,roy2016monocular,li2018megadepth,fu2018deep,godard2019digging,zhao2020towards,Ranftl2022,depthanything,depth_anything_v2,wang2024moge} in the past decade. While most of them learn monocular depth with an error that is invariant to scale or affine transformation, thanks to large-scale data and vision foundation models in recent years, researchers have also developed models \cite{bhat2021adabins, yin2023metric3d, bhat2023zoedepth, li2024binsformer, piccinelli2024unidepth, depthanything, depth_anything_v2,bochkovskii2024depth} that aim to predict absolute (metric) depth from images. Recently, there have also been works \cite{saxena2024surprising,zeng2024wordepth,marigold,patni2024ecodepth,fu2024geowizard} that demonstrate the benefits of repurposing generative models for monocular depth estimation. With such rapid and active developments, the latest state-of-the-art models~\cite{depth_anything_v2, wang2024moge} have started to show foundation model level performance on open-domain images. This raises increasing attention and demand on studying how the off-the-shelf monocular depth models can best benefit geometric vision tasks. Our proposed method works with any monocular depth models and achieves improvement on relative pose estimation with both relative and metric depth priors.

\noindent\textbf{Relative Pose Estimation:} 
Classic relative pose estimators mainly rely on cross-image point correspondences, which either comes from the detection \cite{lowe2004distinctive,detone18superpoint,dusmanu2019d2,tyszkiewicz2020disk,zhao2023aliked} and matching \cite{sarlin20superglue,lindenberger2023lightglue,barath2025stereoglue} of salient keypoints, or direct dense / semi-dense matching \cite{sun2021loftr,wang2022matchformer,edstedt2023dkm,truong2023pdc,edstedt2024roma}. While minimal solvers for point correspondences have been well studied in the two-view case under calibrated~\cite{Nister03FivePt} and uncalibrated conditions~\cite{hartley_zisserman_2004,Stewenius2008}, researchers have been making continuous efforts on improving relative pose estimation with the help of data-driven machine learning, including differentiable robust estimation \cite{ranftl2018deep,wei2023generalized}, relative pose regression \cite{zhou2020learn}, and advanced proposal scoring \cite{barroso2023two}. However, they have not yet achieved the same generalization ability as classic methods in practice, with the LO-RANSAC \cite{chum2003locally,lebeda2012fixing} in PoseLib \cite{PoseLib} remaining the best practical solutions. Very recently, DUSt3R \cite{wang2024dust3r} introduced a 3D foundation model for two-view geometry that benefits from large-scale training, which is further improved by MASt3R \cite{mast3r_arxiv24} with better matching. In this paper, we present a relative pose estimator that consistently improves over PoseLib \cite{PoseLib} with off-the-shelf monocular depth models, and further show that the improvements are orthogonal and can be combined with the recent advances from MASt3R \cite{mast3r_arxiv24}.
\\

\noindent\textbf{Monocular Depth Priors for Geometric Tasks:}
As monocular depth estimation become increasingly robust, there has been growing interests on using the dense depth predictions to benefit downstream geometric tasks, including visual odometry/SLAM \cite{tateno2017cnn,czarnowski2020deepfactors, loo2021deeprelativefusion,studiosfm}, dense reconstruction \cite{pizzoli2014remode,luo2020consistent,wei2021nerfingmvs,yu2022monosdf}, and novel view synthesis \cite{song2023darf,xu2024depthsplat}. However, only a few works have attempted employing monocular depth priors to enhance relative pose estimation. Barath \textit{et al.}~\cite{barath2022relative} propose a 2pt+D solver and also attempt to incorporate affine correspondence \cite{eichhardt2020relative}. However, the 2-point solver is intrinsically degenerate for rigid alignment due to rank deficiency. In~\cite{studiosfm}, PnP-RANSAC is directly applied on top of monocular depths for small-baseline image pairs. Most recently, Ding \textit{et al.}~\cite{ding2024fundamental} develop fundamental matrix solvers for relative depths, and fCOP \cite{zhang2024fcop} relates monocular depth estimation with canonical object priors for focal length estimation. While all prior works assume the depth priors are up to scale, we show that modeling additional shifts are extremely beneficial in practice, even for off-the-shelf metric depth models. 

%% file: sec/3_problem_formulation.tex
\input{figures/pipeline_overview}
\section{Problem Formulation}

\label{sec:problem_formulation}

Given two images $I_1$ and $I_2$, relative pose estimation aims to estimate the rotation $\bm{R}$ and translation $\bm{t}$ transforming the coordinate frame of the first camera to the second one. The classic approach to solve this problem is to establish a set of corresponding pixel matches between the two images through either sparse feature matching (\textit{e.g.}, SuperPoint~\cite{detone18superpoint} + LightGlue~\cite{lindenberger2023lightglue}) or dense correspondence techniques (\textit{e.g.}, RoMa~\cite{edstedt2024roma}), then solve for the essential matrix relating these point matches.

In addition to the 2D correspondences, we aim to leverage depth maps $D_1$ and $D_2$ predicted by some off-the-shelf monocular depth estimation (MDE) model for each image. These depth maps provide valuable 3D priors but come with inherent ambiguities due to the nature of monocular depth estimation. State-of-the-art MDE models~\cite{eftekhar2021omnidata, birkl2023midas3.1, marigold, depth_anything_v2} are often trained with scale- and shift-invariant loss~\cite{Ranftl2022}. They predict either disparity~\cite{Ranftl2022, birkl2023midas3.1, depth_anything_v2} or relative depth~\cite{eftekhar2021omnidata, marigold, depth_anything_v2} that are related to the actual value with an unknown linear (affine) transformation. In our formulation, we focus on affine-invariant relative depth and model this transformation as:
\begin{equation}
    \label{eq:affine-depth}
    \widehat{D}_i = a_iD_i + b_i, \mspace{15mu} i \in \{1, 2\}
\end{equation}
where $a_i$ and $b_i$ are unknown scalar values of the scale and shift of both depth maps.

For a pair of image correspondence $(\textbf{p}_{1j}, \textbf{p}_{2j})$, and their corresponding predicted depth values after the unknown linear transformation $\widehat{d}_{1j}$ and $\widehat{d}_{2j}$, we can lift the 2D points to 3D from both camera views:
\begin{equation}
    \label{eq:lift-point}
    \textbf{P}_{ij} = \bm{K}_i^{-1}
    \begin{bmatrix}
        \textbf{p}_{ij} \\ 1
    \end{bmatrix}\cdot \widehat{d}_{ij}, \mspace{15mu} i \in \{1, 2\}
\end{equation}
where $\bm{K}_i$ is the camera intrinsic matrix of the $i$-th image.

The two 3D points $\textbf{P}_{1j}$ and $\textbf{P}_{2j}$ are related by the rotation $\bm{R}$ and translation $\bm{t}$ of the relative pose:
\begin{equation}
    \label{eq:transform-lifted}
    \textbf{P}_{2j} = \bm{R}\textbf{P}_{1j} + \bm{t}.
\end{equation}
Thus, the monocular depth maps could provide additional constraints on the relative pose.
However, explicitly modeling the scales and shifts for both views is necessary to obtain such constraint. In relative pose estimation, the actual scale of the scene is inherently ambiguous, so estimating the relative scale ratio $\alpha = a_2 / a_1$ is sufficient. Therefore, in addition to the relative rotation $\bm{R}$ and translation $\bm{t}$, we also explicitly estimate two scalar shifts $\beta_1 = b_1 / a_1$, $\beta_2 = b_2 / a_2$, and one scale $\alpha$, which transforms the two depth maps:
\begin{equation}
\label{eq:depth-scale-shift}
    \widehat{D}_1 = D_1 + \beta_1, \mspace{15mu} \widehat{D}_2 = \alpha(D_2 + \beta_2).
\end{equation}

While this formulation assumes the depth priors are affine-invariant relative depth, using our method with metric depth predictions can also lead to more accurate and robust relative pose as we will show empirically in the experiments. This formulation is also illustrated in~\cref{fig:formulation}.

%% file: figures/pipeline_overview.tex
\begin{figure*}[t]
    \centering
    \includegraphics[width=0.95\textwidth, height=0.34\textwidth]{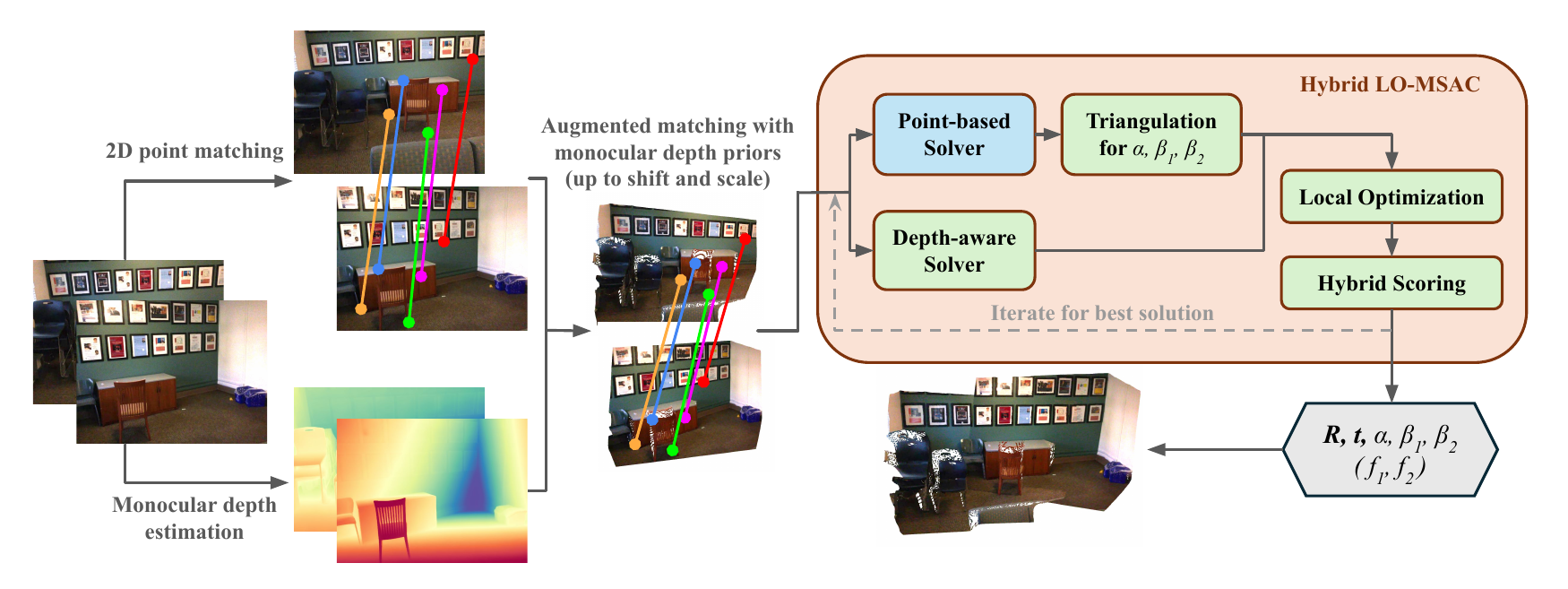}
    \caption{\textbf{Pipeline overview:} Our method takes a pair of images as input, runs off-the-shelf feature matching and monocular depth estimation, then jointly estimates the relative pose, scale and shift parameters of the two depth maps, and optionally the focal lengths. Our method incorporates monocular depth priors in all stages (in green) of hybrid LO-MSAC~\cite{hybridransac, LOMSAC}, including 3 new depth-aware solvers, while still being able to leverage traditional point-based solvers~\cite{Nister03FivePt,hartley_zisserman_2004,Stewenius2008} (in blue).}
    \label{fig:pipeline}
\end{figure*}

%% file: sec/4_method.tex
\section{Method}
\label{sec:method}
We propose three solvers that solve for the relative pose $\bm{R}, \bm{t}$ as well as the shifts $\beta_1, \beta_2$ and scale $\alpha$ using the feature correspondences and depth priors:
\begin{enumerate}
    \item Calibrated solver when all camera intrinsics are known;
    \item Shared-focal solver when both images are captured by camera with a shared unknown focal length;
    \item Two-focal solver when focal lengths are unknown for both cameras.
\end{enumerate} 

The solvers work in a 2-step manner. First, find all possible values for the depth scale and shifts $\alpha, \beta_1, \beta_2$, as well as the focal lengths if unknown. Then, solve for the rotation and translation for each set of the possible scales and shifts. 

\subsection{Solving Depth Scale, Shifts, and Focal Lengths}
\label{sec:solvers}

Given a sampled set of $M$ pixel correspondences $\{(\textbf{p}_{11}, \textbf{p}_{21}), \dots, (\textbf{p}_{1M}, \textbf{p}_{2M})\}$, and the corresponding depth priors from the predicted depth maps $d_{ij} = D_i(\textbf{p}_{ij})$, the solvers rely on the assumption that the corresponding pair-wise distances $\bm{\delta}$ among the lifted 3D points should be equal in the two camera frames. Formally, let us define
\begin{equation*}
    \bm{\delta}^{(i)}_{jk} = \textbf{P}_{ij} - \textbf{P}_{ik}, \text{where}\mspace{10mu} i \in \{1, 2\} \mspace{10mu} \text{and} \mspace{10mu} j \neq k.
\end{equation*}
Because the rotation and translation transformation in \eqref{eq:transform-lifted} is length-preserving, the following equality constraint must hold for all pairs $(j,k)$ of the 3D points:
\begin{equation}
    \label{eq:scale-shift-constraint}
    \norm{\bm{\delta}^{(1)}_{jk}}^2 = \norm{\bm{\delta}^{(2)}_{jk}}^2.
\end{equation}

We can thus obtain a system of $\binom{M}{2}$ equations where the rotation and translation have been eliminated. The three solvers require different numbers of correspondences to solve these equations due to the different numbers of unknowns. In addition to the three values $\alpha, \beta_1, \beta_2$, the shared-focal solver also solves for the shared focal $f$, and the two-focal solver solves for the two focals $f_1, f_2$. \\

\noindent\textbf{Calibrated Solver:}
The calibrated solver addresses the scenario where both cameras have known intrinsic parameters. 
In this case the problem is minimal with $M=3$ points (along with their associated depth priors), as we get ${3 \choose 2} = 3$ equations in \eqref{eq:scale-shift-constraint} in the three unknowns $\alpha, \beta_1$ and $\beta_2$.

Since the intrinsics $\bm{K}_1$ and $\bm{K}_2$ are known, let $\widetilde{\textbf{P}}_{ij} = \bm{K}_i^{-1}\begin{bmatrix}\textbf{p}_{ij}\\ 1\end{bmatrix} = (\widetilde{x}_{ij}, \widetilde{y}_{ij}, 1)^\top$. We have:
\begin{equation*}
\footnotesize
\begin{aligned}
    \norm{\bm{\delta}^{(i)}_{jk}}^2 &= \norm{\widehat{d}_{ij}\widetilde{\textbf{P}}_{ij} - \widehat{d}_{ik}\widetilde{\textbf{P}}_{ik}}^2 \\
    &= \norm{\widehat{d}_{ij}(\widetilde{x}_{ij}, \widetilde{y}_{ij}, 1)^\top - \widehat{d}_{ik}(\widetilde{x}_{ik}, \widetilde{y}_{ik}, 1)^\top}^2 \\
    &= (\widehat{d}_{ij}\widetilde{x}_{ij} - \widehat{d}_{ik}\widetilde{x}_{ik})^2 + (\widehat{d}_{ij}\widetilde{y}_{ij} - \widehat{d}_{ik}\widetilde{y}_{ik})^2 + (\widehat{d}_{ij} - \widehat{d}_{ik})^2.
     \\
\end{aligned}
\end{equation*}
This yields three quartic equations of $\alpha, \beta_1$ and $\beta_2$.
In the equations, $\alpha$ only appears as squares, so we re-parameterize with $\gamma = \alpha^2$ to make the equations cubic and reduce the number of solutions.
Using the method from Larsson \textit{et al.}~\cite{larsson2017efficient} we find that the system of polynomials have at most 4 solutions and create a Gröbner basis solver.
The solver performs linear elimination on a 12x12 matrix followed by a 4x4 eigenvalue problem. \\

\noindent\textbf{Shared-Focal Solver:}
The shared-focal solver is designed for scenarios where both images are captured by camera with an unknown shared focal length $f$. This situation is common in applications such as visual odometry or structure-from-motion with a single moving camera. 
We assume that the focal length is the only unknown intrinsic parameter, and thus we can assume without loss of generality that $\bm{K} = \text{diag}(f,f,1)$. We then have
\begin{equation} \label{eq:p_ij_calib}
\widetilde{\textbf{P}}_{ij} = \bm{K}^{-1}\begin{bmatrix}\textbf{p}_{ij}\\ 1\end{bmatrix} = (\frac{{x}_{ij}}{f}, \frac{{y}_{ij}}{f}, 1)^\top.    
\end{equation}
As we now have an additional unknown, one more correspondence is needed to solve for $\alpha, \beta_1, \beta_2$, and $f$.
However, taking all equations \eqref{eq:scale-shift-constraint} from $M=4$ matches will give an overdetermined system.
We therefore only select 4 out of the 6 possible equations from inserting \eqref{eq:p_ij_calib} into \eqref{eq:scale-shift-constraint}.
Similar to the $\alpha$ reparametrization in the calibrated case, we additionally perform substitution $\omega = 1/f^2$.
After these two substitutions we get four equations of degree 4.
Again applying the method from Larsson \textit{et al.}~\cite{larsson2017efficient} we find that the problem has 8 solutions and we generate a solver with template size 36x36. \\

\noindent\textbf{Two-Focal Solver:}
Last but not least, the two-focal solver addresses the most general case where the two images are captured by cameras with potentially different, unknown focal lengths.
This scenario is particularly relevant for applications involving crowd-sourced imagery.

This case can also be solved with $M=4$ correspondences, but now requires using 5 out of the 6 equations they generate.
The derivation is similar to the shared focal length case, except that we now have independent calibration matrices $\bm{K}_1 = \text{diag}(f_1,f_1,1)$ and $\bm{K}_2 = \text{diag}(f_2,f_2,1)$ for each camera.
In the equations we now introduce $\omega_1 = 1/f_1^2$ and $\omega_2 = 1/f_2^2$.
Applying the method from Larsson \textit{et al.}~\cite{larsson2017efficient} we surprisingly find that the problem only has 4 solutions, fewer than in the shared focal length case.
The solver however requires linear elimination on a 40x40 matrix followed by solving a 4x4 eigenvalue problem.

\subsection{Finding Rotation and Translation}
Only real solutions of $\alpha, \beta_1, \beta_2$ that result in positive depth values in \eqref{eq:depth-scale-shift} are kept as valid solutions (and similarly for the focal lengths). For each possible solution, we estimate a rigid transformation (a rotation $\bm{R}$ and a translation $\bm{t}$), which can be easily retrieved using the SVD method~\cite{procrustes} aligning the back-projected 3D points from both views which satisfy \eqref{eq:transform-lifted}.
Since relative poses are inherently up to scale, $\bm{R}$ and $\bm{t}$ are the final outputs of our proposed solvers.

\subsection{Evaluating Solutions}
The solutions $\Theta = (\bm{R}, \bm{t}, \alpha, \beta_1, \beta_2)$ (optionally also include $f$ or $f_1, f_2$) need to be evaluated on all correspondences with their depth priors. To do so, we compute the \textit{depth-induced reprojection errors}:
\begin{equation}
\label{eq:E_r}
\begin{aligned}
    E_{r(1\rightarrow2)}(\Theta) &= \norm{\mathbf{\Pi}_2(\bm{R}\textbf{P}_1 + \bm{t}) - \textbf{p}_2}^2 \\
    E_{r(2\rightarrow1)}(\Theta) &= \norm{\mathbf{\Pi}_1(\bm{R}^{-1}\textbf{P}_2 - R^{-1}\bm{t}) - \textbf{p}_1}^2,
\end{aligned}
\end{equation}
where $\textbf{P}_1, \textbf{P}_2$ are the lifted 3D points of a pair of correspondences using the shift-and-scaled depth priors $\widehat{d}_1 = d_1 + \beta_1, \widehat{d}_2 = \alpha(d_2 + \beta_2)$, and $\mathbf{\Pi}_i(\textbf{P})$ is the camera projection using $\bm{K}_i$. $\bm{K}_i$ is either known with calibrated cameras, or assembled using the estimated focal lengths solved by solvers. Note that the errors are computed separately for two directions. This is designed to be robust against inconsistencies between the depth maps when combining into a MSAC score with a squared inlier threshold $\tau_r$:
\begin{equation}
    \overline{E}_r(\Theta) = \min(E_{r(1\rightarrow2)}, \tau_r) + \min(E_{r(2\rightarrow1)}, \tau_r).
\end{equation}

\subsection{Hybrid Estimation}
While monocular depth estimation models can provide useful geometric cues, solely relying on the depth priors can be prone to erroneous results when the priors are unreliable. We further propose to jointly use our depth-aware solvers and reprojection errors together with the classic point-based solvers and epipolar (Sampson) error $E_s$~\cite{Sampson1982FittingCS,Luong1996TheFM} in a hybrid LO-MSAC~\cite{hybridransac, LOMSAC} framework. For the calibrated case, the Sampson error is rescaled with the focal lengths to ensure both errors are in pixels.

This approach enhances the estimation with good depth priors but can also utilize 2D correspondences when the priors are unsatisfactory. As shown in \cref{fig:pipeline}, each iteration of the hybrid LO-MSAC pipeline starts by selecting a solver between a depth-aware and a point-based solver depending on the setting:
\begin{itemize}
    \item \textbf{Calibrated}: calibrated solver \& 5-pt essential matrix solver~\cite{Nister03FivePt};
    \item \textbf{Shared-focal}: shared-focal solver \& 6-pt shared focal relative pose solver~\cite{Stewenius2008};
    \item \textbf{Two-focal}: two-focal solver \& 7-pt fundamental matrix solver~\cite{hartley_zisserman_2004}.
\end{itemize}
For point-based solvers, we also need to estimate $\alpha, \beta_1, \beta_2$ consistent with the estimated pose and potentially focal lengths. This is done by triangulating 3D points from the $M$ correspondences, and fitting the depth priors to the projected depth to find the scale and shift in least-squares manner. For the 7-pt fundamental matrix solver, the possible focal lengths are retrieved using Bougnoux formula\cite{bougnoux1998projective}.

The two solvers are initially randomly selected with equal probabilities, and are later chosen using the inlier ratios of the data types following~\cite{hybridransac}. Specifically, we view each correspondences as three different data types: $(\textbf{p}_1, \textbf{p}_2, d_1)$, $(\textbf{p}_1, \textbf{p}_2, d_2)$, and $(\textbf{p}_1, \textbf{p}_2)$. The first two are evaluated using ${E}_{r(1\rightarrow2)}$ and ${E}_{r(2\rightarrow1)}$ respectively, and the third using $E_s$. Inliers are defined as correspondences (in each data type) with error within the squared thresholds: $E_r < \tau_r$ or $E_s < \tau_s$. The local optimization runs on the inliers with least square optimizer that jointly optimizes all parameters of $\Theta$ to minimize the cost function
\begin{equation}
    E(\Theta) = \sum_{i\in I_1}E_{r(1\rightarrow2)} + \sum_{i\in I_2}E_{r(2\rightarrow1)} + 2\lambda_s\frac{\tau_r}{\tau_s}\sum_{i\in I_3}E_s.
\end{equation}
with $I_1, I_2, I_3$ being the inlier sets of the three data types, and $\lambda_s$ is the Sampson error weight (tunable, empirically set to 1 in our experiments).

With $\overline{E}_s = \min(E_s, \tau_s)$, the combined hybrid MSAC score of all correspondences is then
\begin{equation}
    \overline{E}(\Theta; \textbf{p}_1, \textbf{p}_2) = \overline{E}_r + 2\lambda_s\frac{\tau_r}{\tau_s}\overline{E}_s,
\end{equation}
which is used to select the best solution with minimal $\overline{E}$ summed over all correspondences.

%% file: sec/5_exp.tex
\section{Experiments}
\label{sec:exp}

\subsection{Experimental Setup}
\label{sec:exp-setup}
\textbf{Datasets}: We evaluate our method across multiple datasets: ScanNet-1500~\cite{dai2017scannet} and MegaDepth-1500~\cite{li2018megadepth} are used to evaluate relative pose estimation performance with calibrated cameras on indoor and outdoor images. 
ScanNet-1500 is also used for the unknown shared-focal case, and MegaDepth-1500 for the two unknown focal case. To also evaluate the relative pose estimation with unknown, different focal lengths on indoor images, we sample 1064 image pairs from the Stanford 2D-3D-S panoramic dataset~\cite{2d3ds} with random focal lengths using the procedure detailed in Supp. Sec. A. 
In addition, we also sampled 1451 image pairs from the 7 indoor scenes\footnote{ETH3D indoor scenes: \texttt{delivery\_area, kicker, office, pipes, relief, relief\_2, terrains}.} of ETH3D~\cite{ETH3D} dataset. These include all covisible image pairs with a minimum of 50 covisible ground-truth (GT) 3D points. Having these GT 3D points allows us to analyze the performance of our framework especially in hard cases where the image pair has a low covisibility (\cref{fig:covis_eth3d}). \\

\noindent\textbf{Baselines}: We compare against the three classic point-based methods for the three settings.
We use the PoseLib~\cite{PoseLib} RANSAC implementation of each point-based baseline. We also compare with previous methods in their corresponding settings: the Perspective-n-Points (PnP) method~\cite{FischlerB81} used in~\cite{studiosfm} and the solvers proposed in~\cite{barath2022relative} for the calibrated setting; the 3p3d solver~\cite{ding2024fundamental} for the unknown shared-focal setting; and the 4p4d solver~\cite{ding2024fundamental} for the two-focal setting. The solvers from~\cite{barath2022relative} and~\cite{ding2024fundamental} are run using the implementation obtained from the authors within GC-RANSAC~\cite{GCRansac2018} as in the original papers. We also compare with two recent two-view dense reconstruction methods: DUSt3R~\cite{wang2024dust3r} and MASt3R~\cite{mast3r_arxiv24}, in the shared-focal and two-focal settings. DUSt3R and MASt3R are evaluated using the direct output two-view pointmaps from the models without global alignment~\cite{wang2024dust3r}. \\

\noindent\textbf{Feature Matchers and MDE Models}: Our framework can work with off-the-shelf feature matchers and MDE models that produce affine-invariant depth priors. We test our method with sparse matchings Superpoint \cite{detone18superpoint}+LightGlue \cite{lindenberger2023lightglue}, as well as the state-of-the-art (SOTA) dense matcher RoMa \cite{edstedt2024roma}. Several MDE models predicting affine-invariant depths are used: Omnidata \cite{kar20223d-omnidata}, Marigold \cite{marigold}, the metric depth models of Depth-Anything v1 \cite{depthanything} and v2 \cite{depth_anything_v2}, and MoGe \cite{wang2024moge}. Due to limited space, we report results with best performing models in the main paper and put the others in Supp. Sec. B. 
Note that while the Depth-Anything metric models are trained to predict metric depth, \cref{tab:ablat-shift} show that modeling the scale and shift is still beneficial. \\

\noindent\textbf{Implementation Details:} We implement our solvers and hybrid estimation pipeline in C++. Our hybrid LO-MSAC is based on the hybrid-RANSAC implementation from RansacLib \cite{ransaclib}, with non-minimal solver and least square optimization implemented using automatic differentiation from Ceres solver \cite{Agarwal_Ceres_Solver_2022}. We further export Python bindings using pybind11 \cite{pybind11} to conveniently use our framework in Python scripts. For all experiments, ScanNet images are resized to 640x480, ETH3D images to 720x480, and other images are used with their original resolution.

\input{tables/calib_result}
\input{tables/megadepth-calib}

\subsection{Results and Analysis}

\noindent\textbf{Calibrated Cameras: }
\cref{tab:results-calib-scannet} shows the median rotation error $\varepsilon_{\bm{R}}$ and translation error $\varepsilon_{\bm{t}}$, as well as the Area Under the Curve (AUC) of the pose error (max of $\varepsilon_{\bm{R}}$ and $\varepsilon_{\bm{t}}$) for various thresholds on ScanNet-1500~\cite{dai2017scannet}.
With sparse feature matching (SP+SG or SP+LG) as well as dense matching (RoMa), our method consistently improves over the baselines across all metrics using priors from both the Depth-Anything-v1~\cite{depthanything} metric model (DA-met.) and the non-metric affine-invariant depth model MoGe \cite{wang2024moge}. Compared to the PnP baseline, even when the MDE model is trained to predict metric depth, our affine-invariant modeling of the depth priors is still beneficial. Comparisons with other solver combinations from~\cite{barath2022relative} can be found in Supp. Sec. B.

We also evaluate on the outdoor image pairs of MegaDepth-1500 \cite{li2018megadepth}. Due to the larger and more varied scene scales, MDE models often perform poorly on outdoor data compared to indoor. However, as the results in~\cref{tab:megadepth-calib} show, our method can still improve over the baselines on outdoor images with the latest MoGe \cite{wang2024moge} model. While DA-met. depth estimations becomes less reliable outdoor, our method still outperforms RANSAC PnP using the same depth priors and is reasonably close to the point-based baseline thanks to the hybrid estimation. \\

\input{tables/shared_result}
\noindent\textbf{Shared-focal Cameras:}
For the shared-focal case, we again use the ScanNet-1500 image pairs but ignore the GT intrinsics. We additionally report the median $\varepsilon_f$ of relative focal errors which are computed as $\abs{f - f_{gt}} / f_{gt}$. Tab.~\ref{tab:scannet-shared} shows the results.
Note that the 3p3d solver~\cite{ding2024fundamental} takes disparity values as input, so we use the disparity priors predicted by Depth-Anything-v2~\cite{depth_anything_v2} (as in~\cite{ding2024fundamental}).
Our results are reported with the best performing DA-met. model. MoGe is not used for shared-focal or uncalibrated case because it also estimates focal lengths and can therefore benefit directly from using calibrated hybrid estimation.

With all tested matches, our method improves over the baselines consistently and significantly. 3p3d solver only models the depth priors as scale-invariant but does not consider the shift, so it is sensitive to noise in the priors~\cite{ding2024fundamental} and therefore fails to improve over point-based baseline. The recent works DUSt3R~\cite{wang2024dust3r} and MASt3R~\cite{mast3r_arxiv24} produce good image matching, two-view depth maps, and relative pose by directly regressing aligned point maps from two views, so we report their relative pose estimation results for reference and also evaluate our method with MASt3R matches and depth. By using the shared-focal constraint which they do not model, our method performs better than these two-view methods with RoMa matches and DA-met. depth priors, and can be further improved with better matches and depth from MASt3R. \\

\noindent\textbf{Unknown focal cameras:}
For the most general case with two uncalibrated cameras, we define the focal error as the max of the two relative focal errors of the two cameras and again report the median $\varepsilon_f$.
Significant improvements can be observed in~\cref{tab:results-two-focal} on the uncalibrated image pairs from our sampled 2D-3D-S dataset. Our method achieves less than 50\% in the median errors of the point-based baseline and significant improvements in pose error AUCs.

\input{tables/two_focal_result}
\input{tables/megadepth-twofocal}

Similar to the 3p3d solver, the 4p4d solver~\cite{ding2024fundamental} is tested with Depth Anything v2~\cite{depth_anything_v2} depth priors and again fails to improve over point-based baseline. Using better matches from MASt3R~\cite{mast3r_arxiv24}, our method is able to achieve on-par results using monocular depth priors (Depth-Anything-v2 metric model), with significantly better accuracy on small threshold AUCs. Our method further outperforms MASt3R if using the depth from MASt3R as priors. 

We also evaluate with the uncalibrated camera setting on outdoor MegaDepth-1500~\cite{li2018megadepth} images (\cref{tab:megadepth-two-focal}). Compared with the calibrated results in~\cref{tab:megadepth-calib}, our method (with DA-met. model) now shows significant improvements over the point-based fundamental matrix estimator. \\ 

\noindent\textbf{Limited Covisibility:}
We further analyze the pose errors under different levels of covisibility between image pairs on the indoor ETH3D~\cite{ETH3D} pairs and the results are shown in~\cref{fig:covis_eth3d}. Comparing the gap between our method and point-based baselines, incorporating the depth priors can particularly improve the relative pose estimation results on difficult image pairs with limited covisiblity. \\
\input{figures/covis_eth3d}

\input{figures/vis_samples}
\noindent\textbf{Visual examples:}
In Fig.~\ref{fig:vis_samples} we show visualizations of point clouds reconstructed with relative pose estimated by point-based baseline and ground-truth (GT) depth, and results from our method with scale-and-shifted monocular depth priors. The visualizations demonstrate that by incorporating depth priors and by explicit modeling of the affine transformation, our method is able to find a better pose or to avoid failure cases with better depth alignment. \\

\noindent\textbf{Runtime:}
While the added considerations for depth priors introduces additional computations, our method achieves good performance while maintaining efficiency. By tuning LO and RANSAC parameters to balance performance and speed (with only marginal accuracy loss), our pipeline can achieve the following typical (median) solution times per image pair: 31ms for calibrated case, 65ms for shared-focal case on ScanNet-1500; and 129ms for uncalibrated case on 2D-3D-S image pairs. The numbers are measured on Intel Core i7-10700K CPU, with SP+LG matches. The speed can be further improved by replacing the Ceres auto-differentiated Jacobians with analytic expressions, further tuning, and better multi-threading in the pipeline.
When running on a RTX 3080 GPU, the typical runtime of matchers and MDE models are: SP+LG 16ms/pair; DAv2-met. (ViT-L) 0.16s/image; MoGe (ViT-L) 3ms/image. All models are run with default settings. 



\subsection{Ablation Studies}
\label{sec:model_scale_and_shift}

\noindent\textbf{Modeling scale and shift:}
To further demonstrate the effectiveness of our affine correction modeling, we compare with an ablated scale-only baseline that removes the shift modeling in the calibrated solver and scoring. A visual example is shown in teaser~\cref{fig:formulation}. Without the unknown $\beta_1, \beta_2$, the scale and pose can be together solved by finding the rigid transformation using the 3 point matches and depth priors. For scoring, the reprojection error also removes $\beta_1, \beta_2$ and only lift the 2D points with $\alpha$.

\cref{tab:ablat-shift} shows the comparison with the ablated baselines. The non-hybrid version uses only the depth-aware calibrated solver, and only reprojection errors in LO and scoring. The results show that the shift modeling especially makes a difference for non-metric monocular depth models like Marigold \cite{marigold} and for outdoor images due to the larger and more varied scale. With depth priors targeting metric depth, modeling the shift is again proven beneficial.

\input{tables/ablat_shift}

We further analyze when modeling the shift starts making a difference with a synthetic experiment. We add a shift value ranging from 0 to the median of GT depth for each image and use this shifted GT depth as ``depth priors". We compare the median rotation and translation errors on ScanNet-1500~\cite{dai2017scannet} with SP+SG matches. As shown in \cref{fig:pose_errors_shift_gt}, modeling the shift is beneficial as long as the magnitude of $\beta$ is greater than 10\% of the median depth priors, and our method is not influenced by the increased shift magnitude while the errors of the PnP and scale-only baselines increase. In~\cref{fig:gt-shift} we show the computed shift values by fitting depth priors to GT depth with metric \cite{depthanything} and non-metric \cite{marigold} models on indoor \cite{dai2017scannet} and outdoor \cite{li2018megadepth} images. Shift values larger than 10\% of the median of the priors are common even with metric depth models on indoor data, justifying the gains of our method.

\input{figures/pose_errors_shift_gt}
\input{figures/gt_shift}

\input{tables/ablat_hybrid}
\noindent\textbf{Ablation of hybrid estimation:}
\label{sec:ablat-models}
~\cref{tab:ablat-hybrid} shows ablation on different components of our hybrid estimation. While the hybrid scoring brings the largest improvements, all components contribute to the success of our hybrid estimation.

%% file: tables/calib_result.tex
\begin{table}[t]
\scriptsize
\setlength{\tabcolsep}{3pt}
\centering
\begin{tabular}{ccccc|ccc} 

\toprule
\multirow{2}{*}{Matches} & \multirow{2}{*}{Method} & \multirow{2}{*}{MD Model} & \multicolumn{2}{c}{Med. Err. $\downarrow$} & \multicolumn{3}{c}{Pose Error AUC (\%) $\uparrow$} \\
& & & $\varepsilon_{\bm{R}}(\degree)$ & $\varepsilon_{\bm{t}}$(\degree) & @5\degree & @10\degree & @20\degree \\

\midrule
\multirow{7}{*}{SP+LG} & PoseLib-5pt & - & 1.86 & 5.53 & 21.55 & 39.11 & 55.60 \\
& PoseLib-PnP & DA-met. & 2.03 & 6.44 & 15.05 & 34.16 & 53.97 \\
& PoseLib-PnP & MoGe & 1.72 & 5.37 & 19.69 & 40.21 & 58.73 \\
& 2PT+D\&5pt~\cite{barath2022relative} & DA-met. & 1.90 & 5.67 & 20.62 & 38.45 & 54.94 \\
& 2PT+D\&5pt~\cite{barath2022relative} & MoGe & 1.88 & 5.66 & 20.74 & 38.53 & 54.98 \\
& Ours-calib & DA-met. & \underline{1.68} & \underline{4.97} & \underline{22.41} & \underline{42.18} & \underline{59.96} \\
& Ours-calib & MoGe & \textbf{1.57} & \textbf{4.77} & \textbf{23.36} & \textbf{43.39} & \textbf{61.08} \\

\midrule
\multirow{7}{*}{RoMa} & PoseLib-5pt & - & 1.29 & 3.18 & 33.40 & 55.34 & 72.41 \\
& PoseLib-PnP & DA-met. & 1.63 & 4.46 & 22.83 & 47.04 & 68.12 \\
& PoseLib-PnP & MoGe & 1.38 & 3.75 & 28.38 & 52.62 & 71.61 \\
& 2PT+D\&5pt~\cite{barath2022relative} & DA-met. & 1.32 & 3.27 & 32.27 & 54.34 & 71.72 \\
& 2PT+D\&5pt~\cite{barath2022relative} & MoGe & 1.33 & 3.29 & 32.21 & 54.29 & 71.62 \\
& Ours-calib & DA-met. & \underline{1.26} & \underline{3.14} & \textbf{34.21} & \underline{56.55} & \underline{73.55} \\
& Ours-calib & MoGe & \textbf{1.24} & \textbf{3.12} & \underline{34.26} & \textbf{56.77} & \textbf{73.67} \\

\bottomrule
\end{tabular}
\caption{Results of relative pose estimation with known camera intrinsics on ScanNet-1500~\cite{dai2017scannet}. The reported metrics are median rotation and translation direction errors $\varepsilon_{\bm{R}}$,  $\varepsilon_{\bm{t}}$, as well as pose error AUCs under 5\degree, 10\degree, and 20\degree thresholds. Best results with each type of matches are \textbf{bolded} and second best \underline{underlined}.}
\label{tab:results-calib-scannet}
\end{table}

%% file: tables/megadepth-calib.tex
\begin{table}[t]
\scriptsize
\centering
\setlength{\tabcolsep}{3pt}
\begin{tabular}{ccccc|ccc}

\toprule
\multirow{2}{*}{Matches} & \multirow{2}{*}{Method} & \multirow{2}{*}{MD Model} & \multicolumn{2}{c}{Med. Err. $\downarrow$} & \multicolumn{3}{c}{Pose Error AUC (\%) $\uparrow$} \\
& & & $\varepsilon_{\bm{R}}(\degree)$ & $\varepsilon_{\bm{t}}(\degree)$ & @5\degree & @10\degree & @20\degree \\
\midrule
\multirow{7}{*}{SP+LG} & PoseLib-5pt & - & \underline{0.46} & \underline{1.23} & \underline{60.16} & 74.25 & 84.34 \\
& PoseLib-PnP & DA-met. & 0.99 & 3.29 & 34.79 & 54.55 & 72.32 \\
& PoseLib-PnP & MoGe & 0.56 & 1.80 & 52.46 & 71.05 & 83.80 \\
& 2PT+D\&5pt~\cite{barath2022relative} & DA-met. & 0.52 & 1.37 & 57.83 & 72.85 & 83.73 \\
& 2PT+D\&5pt~\cite{barath2022relative} & MoGe & 0.54 & 1.44 & 55.86 & 70.97 & 82.20 \\
& Ours-calib & DA-met. & 0.47 & 1.26 & 59.80 & \underline{74.77} & \underline{85.47} \\
& Ours-calib & MoGe & \textbf{0.41} & \textbf{1.16} & \textbf{63.48} & \textbf{77.79} & \textbf{87.18} \\
\bottomrule
\end{tabular}
\caption{Relative pose estimation results with known intrinsics on MegaDepth-1500~\cite{li2018megadepth}.}
\label{tab:megadepth-calib}
\end{table}

%% file: tables/shared_result.tex
\begin{table}[bt]
\scriptsize
\centering
\setlength{\tabcolsep}{2pt}
\begin{tabular}{cccccc|ccc} 
\toprule
\multirow{2}{*}{Matches} & \multirow{2}{*}{Method} & \multirow{2}{*}{MD Model} & \multicolumn{3}{c}{Med. Err. $\downarrow$} & \multicolumn{3}{c}{Pose Err. AUC ($\%$) $\uparrow$} \\

& & & $\varepsilon_{\bm{R}}(\degree)$ & $\varepsilon_{\bm{t}}(\degree)$ & $\varepsilon_f$(\%) & @5\degree & @10\degree & @20\degree \\

\midrule
\multirow{3}{*}{SP+LG} & PoseLib-6pt & - & 2.71 & 8.05 & 10.94 & 14.19 & 29.49 & 45.47 \\
& 3p3d~\cite{ding2024fundamental} & DAv2 & 3.27 & 9.37 & 13.66 & 12.09 & 26.09 & 41.88 \\
& Ours-sf & DA-met. & \textbf{1.93} & \textbf{5.78} & \textbf{6.99} & \textbf{17.74} & \textbf{36.71} & \textbf{55.12} \\

\midrule
\multirow{3}{*}{RoMa} & PoseLib-6pt & - & 1.52 & 3.85 & 4.49 & 27.17 & 49.24 & 67.42 \\
& 3p3d~\cite{ding2024fundamental} & DAv2 & 1.64 & 4.15 & 5.30 & 25.46 & 47.17 & 65.47 \\
& Ours-sf & DA-met. & \textbf{1.37} & \textbf{3.54} & \textbf{3.71} & \textbf{29.81} & \textbf{53.11} & \textbf{71.15} \\

\midrule
\multirow{4}{*}{MASt3R} & PoseLib-6pt & - & 1.44 & 3.31 & 4.18 & 30.28 & 54.16 & 72.87 \\
& 3p3d~\cite{ding2024fundamental} & DAv2 & 1.51 & 3.40 & 4.54 & 29.17 & 52.28 & 71.08 \\
& \multirow{2}{*}{Ours-sf} & DA-met. & \underline{1.35} & \underline{3.09} & \textbf{3.73} & \underline{31.87} & \underline{56.20} & \underline{74.51} \\
& & MASt3R & \textbf{1.32} & \textbf{3.06} & \underline{3.79} & \textbf{32.58} & \textbf{56.99} & \textbf{74.91} \\

\midrule
\multicolumn{3}{c}{Reference entry - DUSt3R} & 1.29 & 4.02 & 3.80 & 25.90 & 48.45 & 68.03 \\
\multicolumn{3}{c}{Reference entry - MASt3R} & 1.53 & 4.22 & 5.72 & 23.94 & 46.44 & 66.18 \\
\bottomrule
\end{tabular}

\caption{Uncalibrated shared-focal relative pose estimation results on ScanNet-1500~\cite{dai2017scannet}. In addition to $\varepsilon_{\bm{R}}$ and $\varepsilon_{\bm{t}}$ and pose error AUCs, the median relative focal error $\varepsilon_f$ is also reported. }
\label{tab:scannet-shared}
\end{table}

%% file: tables/two_focal_result.tex
\begin{table}[t]
\scriptsize
\centering
\setlength{\tabcolsep}{1pt}
\begin{tabular}{cccccc|cccc}
\toprule
\multirow{2}{*}{Matches} & \multirow{2}{*}{Method} & \multirow{2}{*}{MD Model} & \multicolumn{3}{c}{Med. Err. $\downarrow$} & \multicolumn{4}{c}{Pose Err. AUC (\%) $\uparrow$} \\
& & & $\varepsilon_{\bm{R}}(\degree)$ & $\varepsilon_{\bm{t}}(\degree)$ & $\varepsilon_f$(\%) & @2\degree & @5\degree & @10\degree & @20\degree \\
\midrule

\multirow{3}{*}{SP+LG} & PoseLib-7pt & - & 16.83 & 21.57 & 59.88 & 5.85 & 13.95 & 21.94 & 30.71 \\
& 4p4d~\cite{ding2024fundamental} & DAv2 & 19.27 & 21.64 & 66.57 & 5.31 & 13.66 & 21.78 & 30.22 \\
& Ours-tf & DAv2-met. & \textbf{5.66} & \textbf{9.26} & \textbf{29.06} & \textbf{9.15} & \textbf{22.22} & \textbf{32.80} & \textbf{43.26} \\

\midrule
\multirow{3}{*}{RoMa} & PoseLib-7pt & - & 7.99 & 8.77 & 37.72 & 8.73 & 20.31 & 30.45 & 41.48 \\
& 4p4d~\cite{ding2024fundamental} & DAv2 & 10.94 & 11.35 & 44.94 & 7.23 & 16.50 & 25.31 & 36.26 \\
& Ours-tf & DAv2-met. & \textbf{3.33} & \textbf{4.32} & \textbf{17.29} & \textbf{13.50} & \textbf{29.19} & \textbf{42.18} & \textbf{54.42} \\

\midrule
\multirow{4}{*}{MASt3R} & PoseLib-7pt & - & 3.21 & 3.34 & 19.95 & 12.58 & 30.27 & 45.57 & 59.85 \\
& 4p4d~\cite{ding2024fundamental} & DAv2 & 3.21 & 3.28 & 20.07 & 12.13 & 29.82 & 45.30 & 60.05 \\
& \multirow{2}{*}{Ours-tf} & DAv2-met. & \underline{1.98} & \underline{2.22} & \underline{11.27} & \underline{18.05} & \underline{39.92} & \underline{56.64} & \underline{70.86} \\
& & MASt3R & \textbf{1.33} & \textbf{1.66} & \textbf{7.93} & \textbf{22.44} & \textbf{48.02} & \textbf{64.79} & \textbf{76.55} \\

\midrule
\multicolumn{3}{c}{Reference entry - DUSt3R} & 2.65 & 4.65 & 7.41 & 6.43 & 24.47 & 42.39 & 58.36 \\
\multicolumn{3}{c}{Reference entry -  MASt3R} & 1.71 & 2.49 & 2.91 & 13.39 & 38.41 & 57.92 & 71.91 \\
\bottomrule
\end{tabular}
\caption{Uncalibrated results on our generated image pairs with different focal lengths from Stanford 2D-3D-S~\cite{2d3ds} dataset. $\varepsilon_f$ is the maximum of the two relative focal errors per image pair.}
\label{tab:results-two-focal}

\end{table}

%% file: tables/megadepth-twofocal.tex
\begin{table}[t]
\scriptsize
\centering
\setlength{\tabcolsep}{4pt}
\begin{tabular}{ccccc|ccc}

\toprule
\multirow{2}{*}{Method} & \multirow{2}{*}{MD Model} & \multicolumn{3}{c}{Med. Err. $\downarrow$} & \multicolumn{3}{c}{Pose Err. AUC (\%) $\uparrow$} \\
& & $\varepsilon_{\bm{R}}(\degree)$ & $\varepsilon_{\bm{t}}(\degree)$ & $\varepsilon_f$(\%) & @5\degree & @10\degree & @20\degree \\
\midrule

PoseLib-7pt & - & 1.97 & 5.72 & 23.64 & 21.23 & 36.80 & 54.89  \\
4p4d~\cite{ding2024fundamental} & DAv2 & 3.48 & 8.70 & 39.92 & 14.04 & 26.70 & 44.08 \\
Ours-tf & DA-met. & \textbf{1.25} & \textbf{4.81} & \textbf{15.99} & \textbf{25.70} & \textbf{42.90} & \textbf{61.79} \\

\bottomrule
\end{tabular}
\caption{Uncalibrated results on MegaDepth-1500~\cite{li2018megadepth}.}
\label{tab:megadepth-two-focal}

\end{table}

%% file: figures/covis_eth3d.tex
\begin{figure}[tb]
    \centering
    \begin{subfigure}[b]{0.48\columnwidth}
        \centering
        \includegraphics[width=\textwidth]{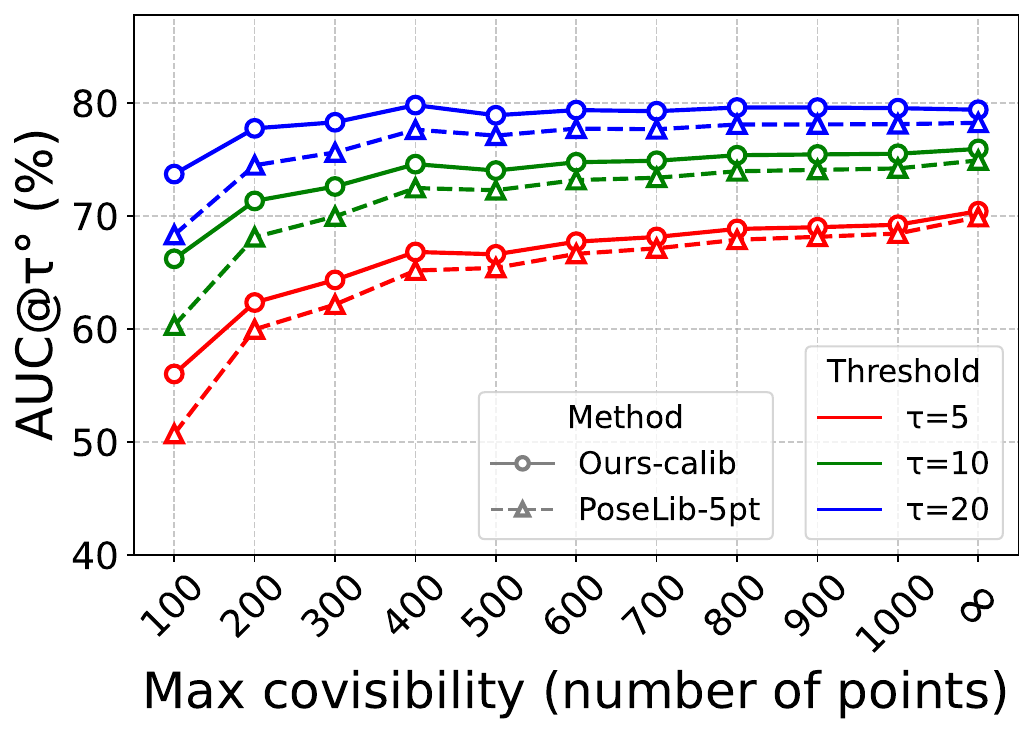}
    \end{subfigure}
    \hspace{4pt}
    \begin{subfigure}[b]{0.48\columnwidth}
        \centering
        \includegraphics[width=\textwidth]{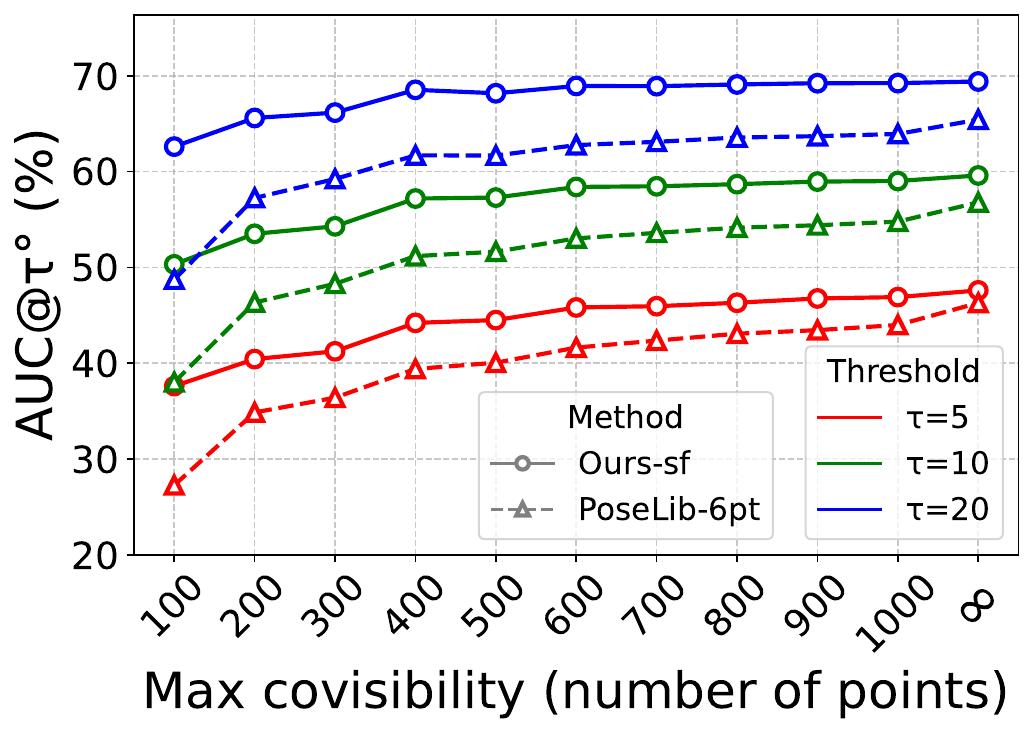}
    \end{subfigure}
    \caption{Pose error AUCs on sampled indoor ETH3D~\cite{ETH3D} image pairs with covisible GT points less than the thresholds on X-axis. \textbf{Left}: calibrated, SP+LG, and MoGe \cite{wang2024moge} priors; \textbf{Right}: shared-focal, SP+LG, and DAv2-met.\cite{depth_anything_v2} metric priors.}
    \label{fig:covis_eth3d}
\end{figure}

%% file: figures/vis_samples.tex
\begin{figure}[tb]
\scriptsize
\centering
\setlength\tabcolsep{1pt} 

\begin{minipage}{\linewidth}
\renewcommand{\arraystretch}{0.4}
\begin{tabular}{cccccc}
\includegraphics[width=0.13\linewidth]{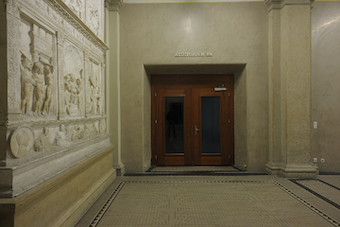} & & &
\raisebox{-0.5\height}[0pt][0pt]{\includegraphics[width=0.27\linewidth]{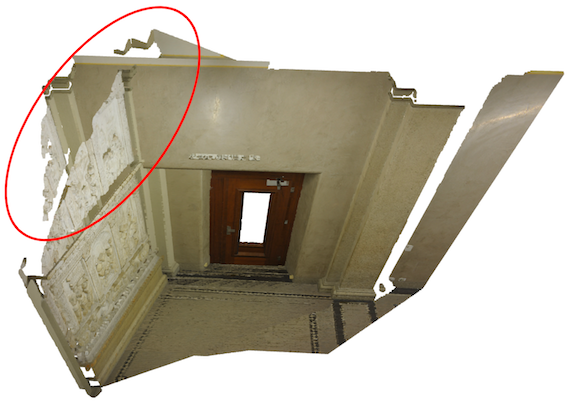}} &
\raisebox{-0.5\height}[0pt][0pt]{\includegraphics[width=0.27\linewidth]{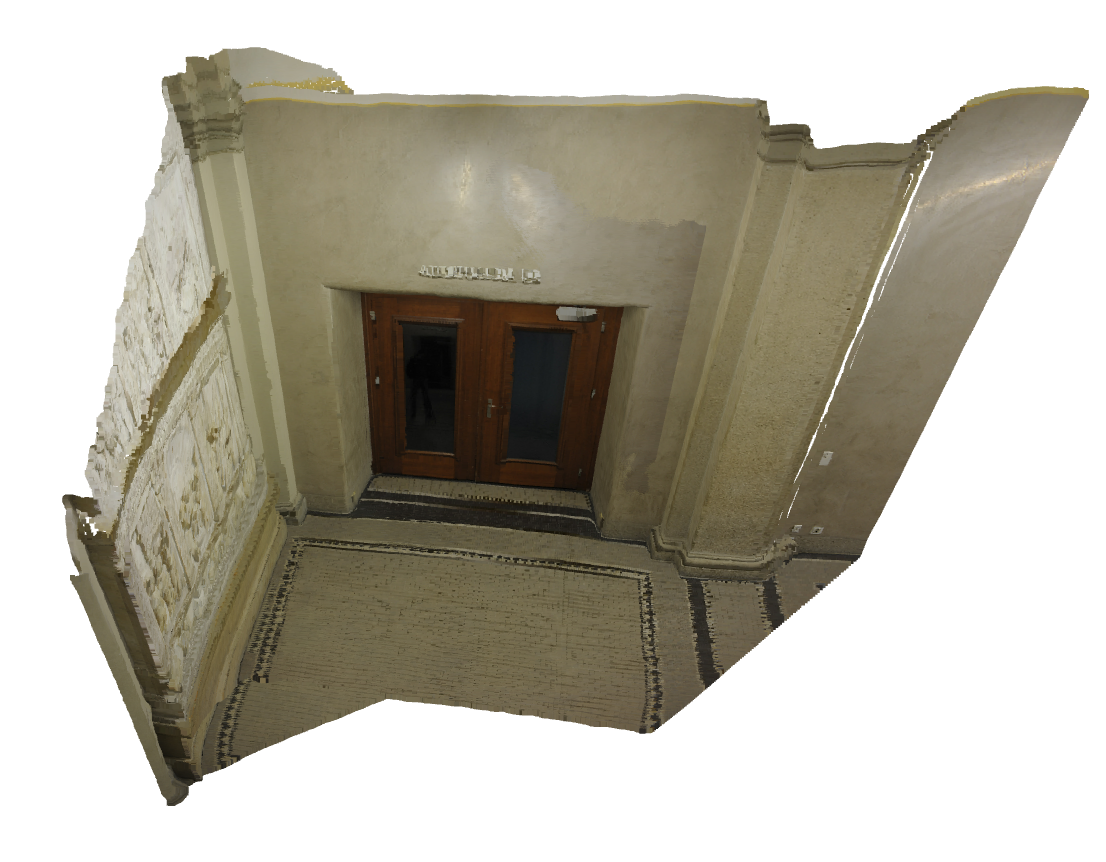}} &
\raisebox{-0.5\height}[0pt][0pt]{\includegraphics[width=0.27\linewidth]{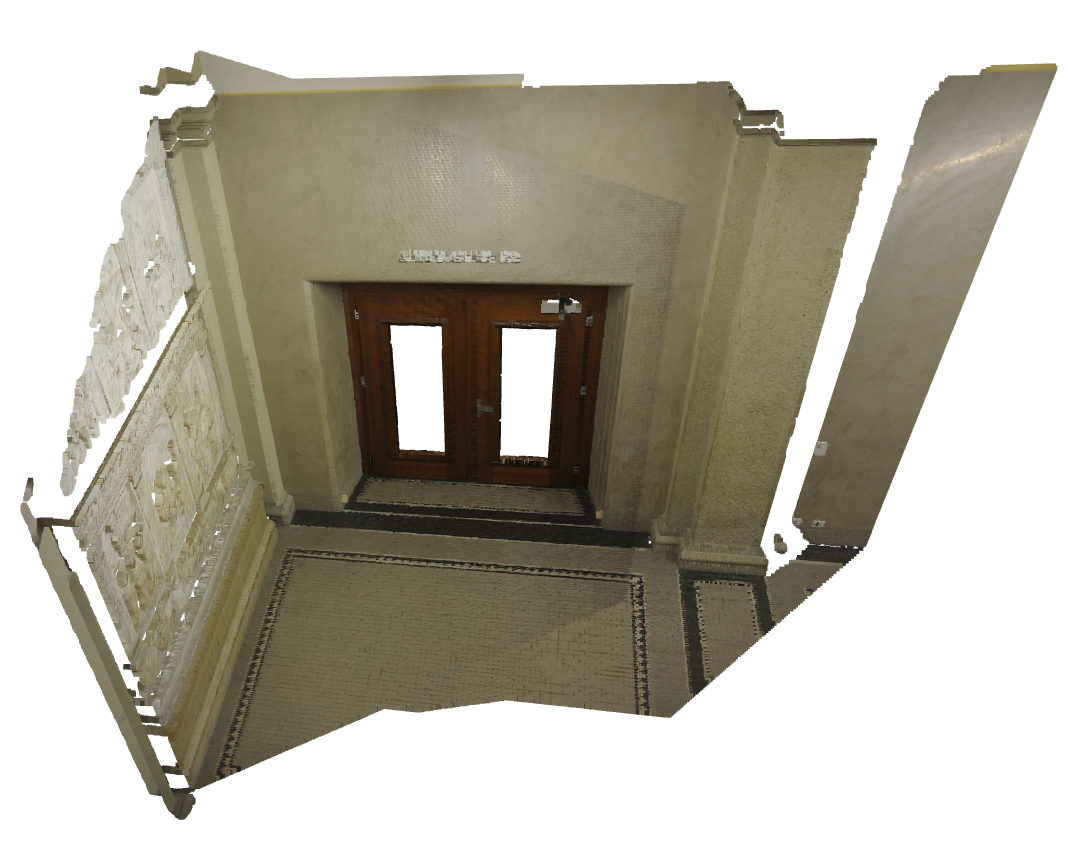}} \\
\includegraphics[width=0.13\linewidth]{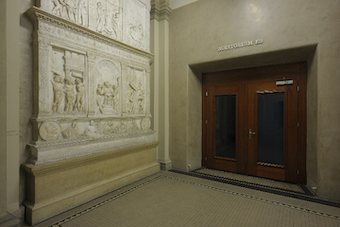} & \\
\includegraphics[width=0.13\linewidth]{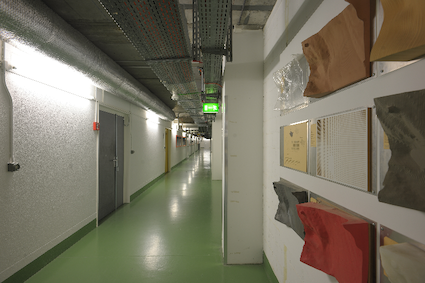} & & &
\raisebox{-0.5\height}[0pt][0pt]{\includegraphics[width=0.27\linewidth]{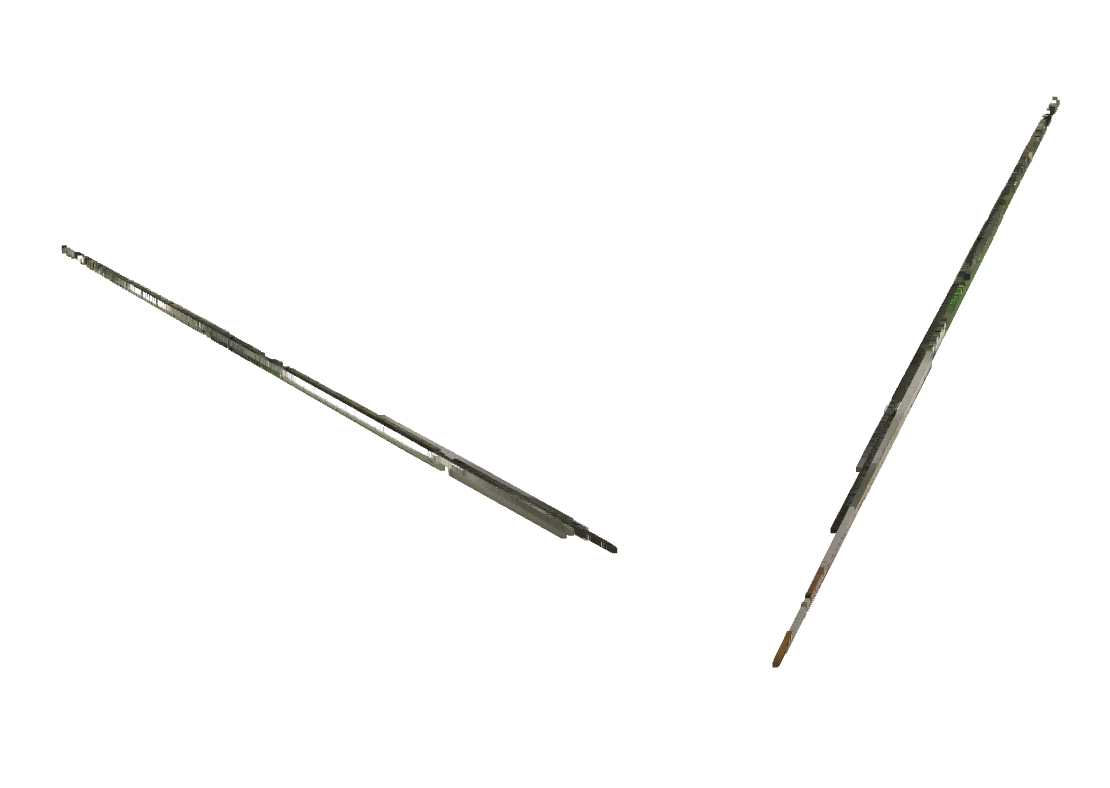}} &
\raisebox{-0.5\height}[0pt][0pt]{\includegraphics[width=0.27\linewidth]{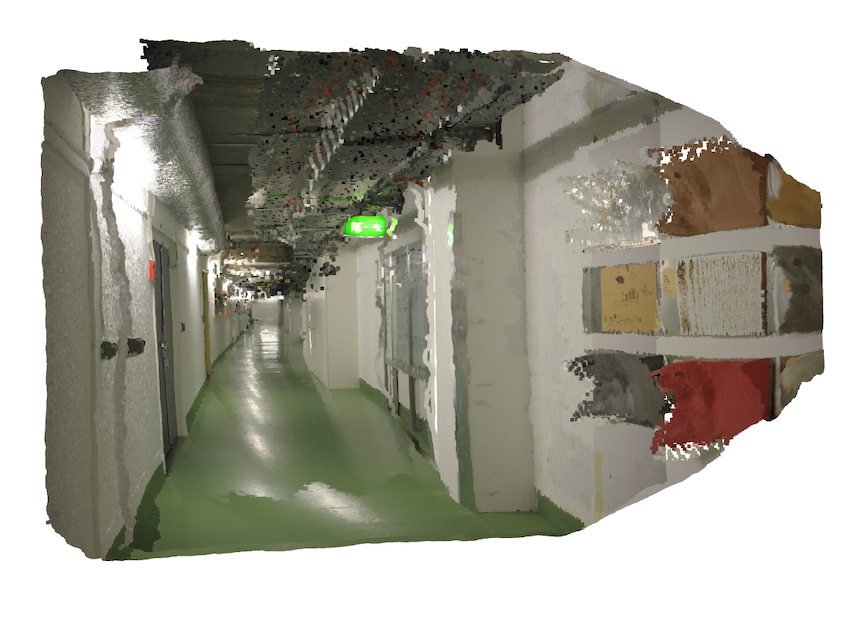}} &
\raisebox{-0.5\height}[0pt][0pt]{\includegraphics[width=0.27\linewidth]{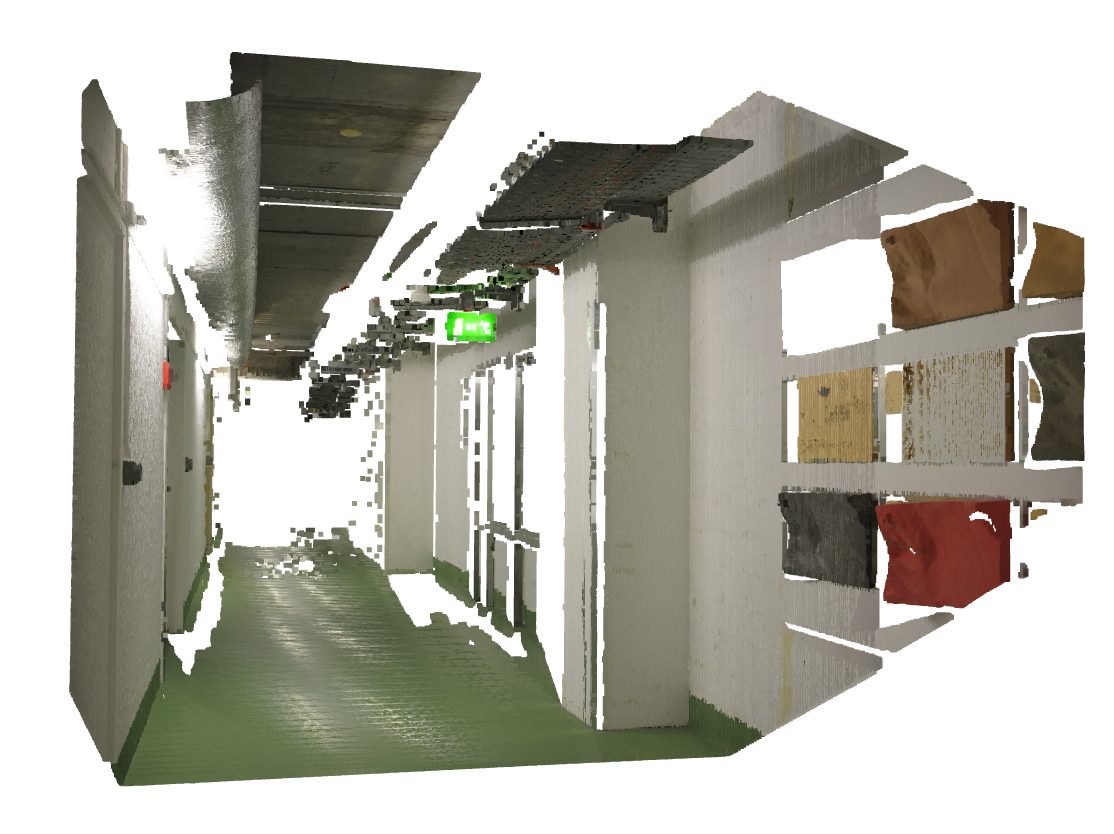}} \\
\includegraphics[width=0.13\linewidth]{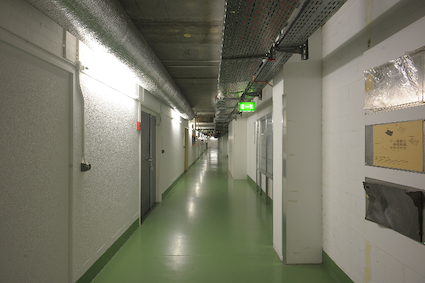} & \\
\end{tabular}
\end{minipage} 
\caption{Visualization on ETH3D~\cite{ETH3D}. \textbf{Left:} back-projected GT depth with pose found by point-based method (translation rescaled to match GT); \textbf{Middle:} back-projected depth priors from Marigold~\cite{marigold} aligned using the scale, shifts, pose, and focal length from our method; \textbf{Right:} GT depth with GT pose.}
\label{fig:vis_samples}
\end{figure}

%% file: tables/ablat_shift.tex
\begin{table}[tb]
\scriptsize
\centering
\setlength{\tabcolsep}{3pt}
\begin{tabular}{cccccc|ccc} 
\toprule
\multirow{2}{*}{Dataset} &
\multirow{2}{*}{MD Model} & \multirow{2}{*}{Hybrid} & \multirow{2}{*}{Shift} & \multicolumn{2}{c}{Med. Err. $\downarrow$} & \multicolumn{3}{c}{Pose Error AUC (\%) $\uparrow$} \\

& & & & $\varepsilon_{\bm{R}}(\degree)$ & $\varepsilon_{\bm{t}}$(\degree) & @5\degree & @10\degree & @20\degree \\
\midrule
\multirow{8}{*}{\makecell{ScanNet \\ -1500}} &
\multirow{4}{*}{DA-met.} & \checkmark & \checkmark & \textbf{1.74} & \textbf{4.73} & \textbf{23.12} & \textbf{43.37} & \textbf{62.26} \\
& & \checkmark & & 1.80 & 5.09 & 19.96 & 41.45 & 61.58 \\
& & & \checkmark & 1.88 & 5.33 & 19.13 & 40.21 & 60.65 \\
& & & & 2.00 & 5.85 & 16.63 & 37.24 & 59.37 \\
\cmidrule{2-9}
& \multirow{4}{*}{Marigold} & \checkmark & \checkmark & \textbf{1.87} & \textbf{5.05} & \textbf{20.68} & \textbf{40.71} & \textbf{59.92} \\
& & \checkmark & & 2.85 & 9.17 & 12.34 & 26.64 & 44.53 \\
& & & \checkmark & 2.19 & 6.41 & 14.70 & 33.73 & 54.75 \\
& & & & 6.24 & 20.44 & 1.09 & 4.84 & 17.15 \\
\midrule

\multirow{4}{*}{\makecell{MegaDepth \\ -1500}} &
\multirow{4}{*}{MoGe} & \checkmark & \checkmark & \textbf{0.42} & \textbf{1.16} & \textbf{63.48} & \textbf{77.44} & \textbf{86.85} \\
& & \checkmark & & 0.63 & 1.90 & 51.32 & 70.09 & 83.02 \\
& & & \checkmark & 0.55 & 1.56 & 56.87 & 73.80 & 85.11 \\
& & & & 0.74 & 2.45 & 44.15 & 64.90 & 79.84 \\
\bottomrule
\end{tabular}
\caption{Ablation of shift modeling with calibrated cameras. All results are with SP+LG correspondences.}
\label{tab:ablat-shift}
\end{table}

%% file: figures/pose_errors_shift_gt.tex
\begin{figure}[tb]
    \centering
    \includegraphics[width=0.9\columnwidth]{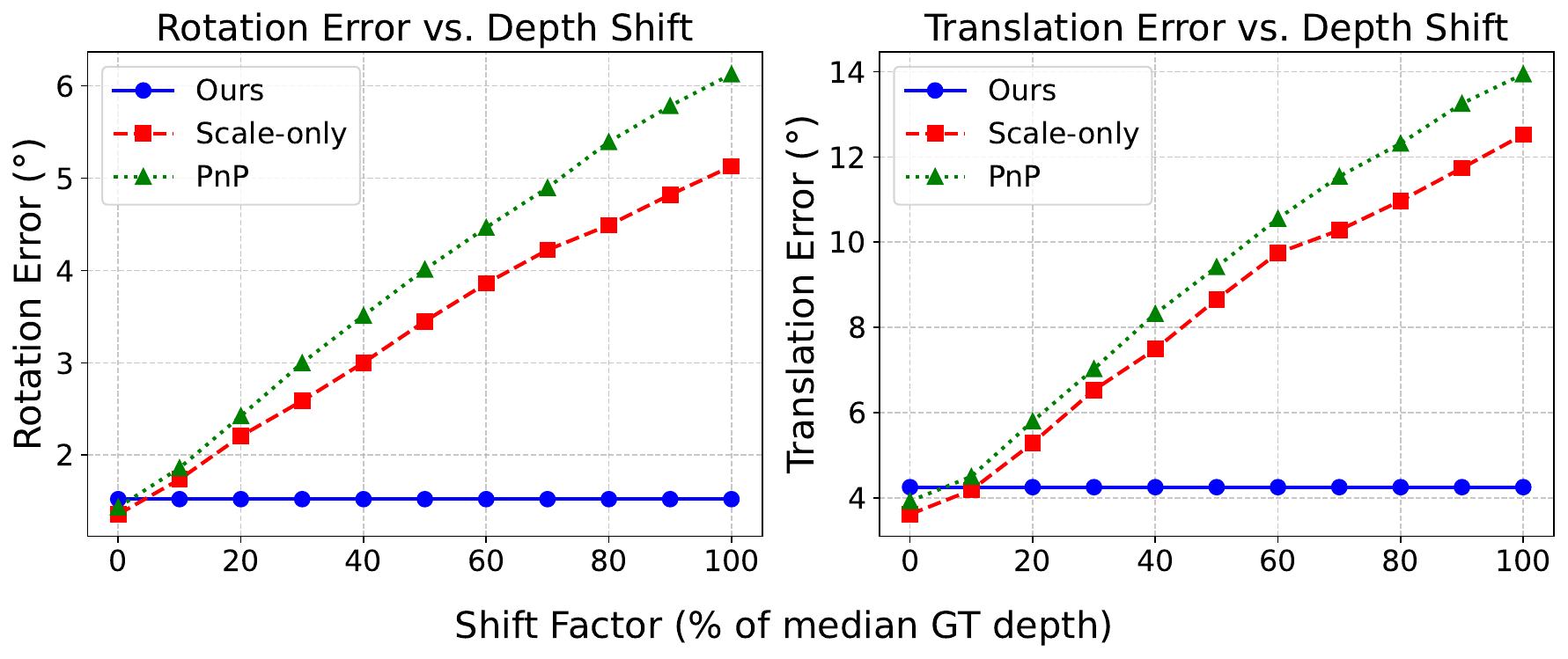}
    \caption{Rotation and translation error by adding shift values to GT depth as "depth priors" on ScanNet-1500 \cite{dai2017scannet}.}
    \label{fig:pose_errors_shift_gt}
\end{figure}

%% file: figures/gt_shift.tex
\begin{figure}[t]
\centering
\scriptsize

\includegraphics[width=0.96\linewidth, height=0.18\linewidth]{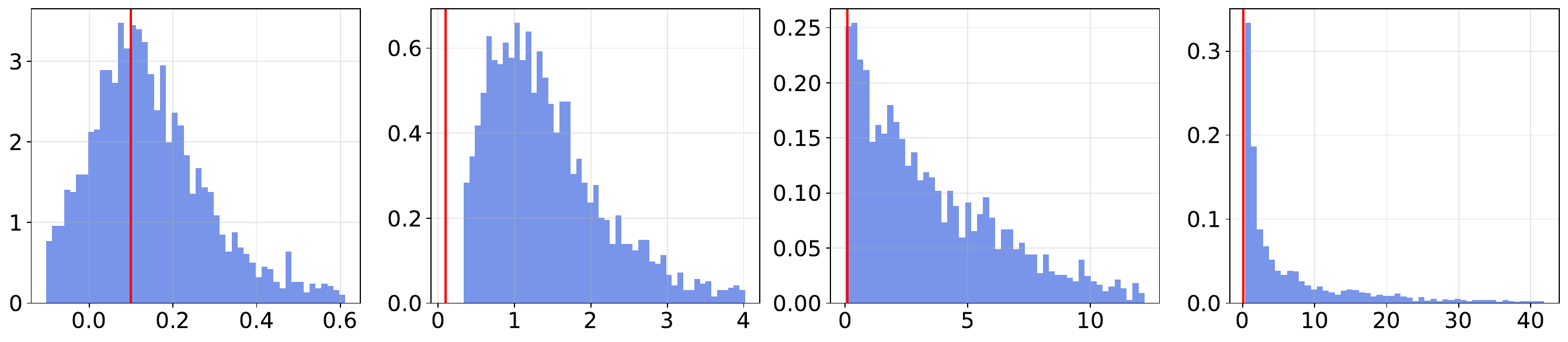}
\begin{tabular}{*{4}{>{\centering\arraybackslash}p{0.2\linewidth}}}
ScanNet & ScanNet & MegaDepth & MegaDepth \\
DA-met. & Marigold & DA-met. & Marigold \\
\end{tabular}
\caption{Distribution (probabilistic density) of GT shift values ($\beta_1, \beta_2$) after fitting MDE depth priors to the GT depth. Red lines indicate $\beta=10\%=0.1$ (refer to~\cref{fig:pose_errors_shift_gt}). Even with metric depth priors on indoor data, modeling the shift is crucial.}
\label{fig:gt-shift}
\end{figure}

%% file: tables/ablat_hybrid.tex
\begin{table}[tb]
\scriptsize
\centering
\setlength{\tabcolsep}{3pt}
\begin{tabular}{cccccc|ccc} 
\toprule

\multirow{2}{*}{Solver} & \multirow{2}{*}{LO} & \multirow{2}{*}{Score} & \multicolumn{3}{c}{Med. Err. $\downarrow$} & \multicolumn{3}{c}{Pose Error AUC (\%) $\uparrow$} \\

& & & $\varepsilon_{\bm{R}}(\degree)$ & $\varepsilon_{\bm{t}}$(\degree) & $\varepsilon_{f}$(\%) & @5\degree & @10\degree & @20\degree \\
\midrule
H & H & H & \textbf{2.00} & \textbf{5.70} & \textbf{6.84} & \textbf{18.35} & \textbf{37.54} & \textbf{57.58} \\
D & H & H & 2.20 & 5.89 & 7.42 & 17.17 & 36.17 & 56.03 \\
P & H & H & 2.20 & 6.10 & 7.36 & 16.60 & 35.33 & 55.26 \\
D & D & H & 2.22 & 6.33 & 7.95 & 15.42 & 34.39 & 55.14 \\
P & P & H & 2.42 & 6.68 & 9.09 & 15.06 & 32.32 & 51.00 \\
D & D & D & 2.21 & 6.83 & 8.18 & 13.85 & 32.11 & 53.50 \\
P & P & P & 2.84 & 7.45 & 11.30 & 14.06 & 29.86 & 48.13 \\

\bottomrule
\end{tabular}
\caption{Ablation of different components of hybrid LO-MSAC on ScanNet-1500~\cite{dai2017scannet} with shared-focal setting, SP+SG matches and DA-met. depth priors. H - Hybrid; D - Depth solver and reprojection score; P - Point-based solver and Sampson score.}
\label{tab:ablat-hybrid}
\end{table}

%% file: sec/6_conclusion.tex
\section{Conclusion}
In this paper, we present a relative pose estimator that benefits from explicit modeling of affine variants of monocular depth priors. We propose solvers for calibrated and uncalibrated camera setups, and combine them with classic methods to achieve consistent improvements across multiple datasets. The proposed method can further benefit from latest advances on image matching and monocular depth estimation. \\

\noindent\textbf{Acknowledgements. } We sincerely thank Philipp Lindenberger for insightful ideas, as well as Lei Li, Chong Bao, Yiming Zhao, and Zihan Zhu for their support and helpful discussions. Viktor Larsson was supported by ELLIIT and the Swedish Research Council (Grant No. 2023-05424).

%% file: sec/X_suppl.tex
\appendix
\maketitlesupplementary

\vspace{\baselineskip}

\section{Additional Details on Experimental Setup}
\label{sec:supp-setup}
\textbf{Generating 2D-3D-S~\cite{2d3ds} image pairs: } As mentioned in~\cref{sec:exp-setup} in the main paper, we hereby provide the details of how the 1064 image pairs from the Stanford 2D-3D-S~\cite{2d3ds} panoramic dataset are generated. 

The dataset includes panoramic scans captured in different rooms across 7 different areas. We generate 4 image pairs from each pair of panoramas captured in a same room by: 
\begin{itemize}
    \item First, generate a random roll ($\pm 30\degree$), pitch ($\pm 30\degree$), yaw ($\pm 180\degree)$, and FOV (60\degree - 105\degree, effectively a random focal length);
    \item Then, project the first image using the generated roll, pitch, yaw, and FOV using equirectangular projection, and crop the image to a randomly chosen crop size among 1080x540, 960x540, 1024x768, 640x480 (WxH);
    \item Next, sample a random pixel within the middle $(W/2, H/2)$ of the first image (thus ensuring reasonable covisibility), and lift-project it into the second panorama using the ground truth depth;
    \item Finally, project the second image using equirectangular projection by centering on this projected point, with again a random roll ($\pm 30\degree$), a random FOV (60\degree - 105\degree), and a random crop size.
\end{itemize} 
Among all generated image pairs, we randomly sample a maximum of 152 image pairs per each of the 7 areas, resulting in a total of 1064 image pairs. \\

\noindent\textbf{Hyperparameters: } The RANSAC thresholds are tuned for best performance for our method as well as the baselines on each dataset. The reprojection error threshold $\tau_r$ for our method is set to 8px for ScanNet~\cite{dai2017scannet} images (resized to 640x480) and ETH3D~\cite{ETH3D} images (resized to 720x480), and 16px for MegaDepth~\cite{li2018megadepth} and Stanford 2D-3D-S~\cite{2d3ds} sampled images. The Sampson error threshold $\tau_s$ for our method is set to 2px on ScanNet and 1px on other datasets. The epipolar error threshold for the PoseLib~\cite{PoseLib} baselines is set to 1px on ETH3D and 2px on other datasets. The threshold of GC-RANSAC~\cite{GCRansac2018} for the solvers from~\cite{ding2024fundamental} is set to 0.75px uniformally. 

As mentioned in the main paper, we empirically fix the Sampson error weight $\lambda_s$ to 1.0 in our experiments to demonstrate the effectiveness of the proposed hybrid estimation. This however can be tunable to adjust to different reliability of the depth priors and feature matchers on different dataset to make the estimator focus more on the depth-augmented correspondences or pure point correspondences.

\section{Additional Experiment Results}
\label{sec:supp-exp}
\input{tables/ablation_monodepth}

\textbf{Results with different monocular depth models:}
Our method is designed to work with any off-the-shelf MDE models, with Depth-Anything variants \cite{depthanything,depth_anything_v2} and MoGe \cite{wang2024moge} giving the best results. The accuracy can further benefit from developments on more accurate MDE models.

We include here a comparison of using different monocular depth estimation (MDE) models with our method across three tasks and three datasets in ~\cref{tab:ablat-monodepth}. Our method can improve upon the baseline with both metric depth priors (Depth-Anything v1~\cite{depthanything} and v2~\cite{depth_anything_v2} metric models) and non-metric relative depth priors (Omnidata~\cite{kar20223d-omnidata}, Marigold~\cite{marigold}, MoGe~\cite{wang2024moge}). MoGe is only evaluated in the calibrated setting due to its ability to also produce a good estimation of focal lengths, and can therefore directly benefit from using the more accurate calibrated estimation in the uncalibrated cases. We include a row using the GT Depth as the ``depth priors" for each task to show the potential of our method with potentially more advanced monocular depth models especially for the shared-focal and two-focal settings. It is worth noting that, while disparity priors in general do not align with our affine-invariant relative depth formulation and inverting those would break the affine-invariance of disparity values, we find that on outdoor images inverting the Depth-Anything-v2~\cite{depth_anything_v2} disparities could lead to better results than its metric depth sibling. We postulate that this is due to the disparity being able to encode a larger range of depths within the output range of the model, which is beneficial for outdoor scenes. \\

\noindent\textbf{Additional comparison with solvers from~\cite{barath2022relative}: }
The solvers proposed in~\cite{barath2022relative} considers monocular depth priors in solving relative poses, thus they are highly related to our work. However, their modeling only considers the scale of the depth priors without the shift. 
We compare our method with the three minimal solver \& non-minimal solver configurations mentioned in~\cite{barath2022relative}: 2PT+D \& 4PT+D, 2PT+D \& 5pt, and Simulated P3P \& 5pt. We use the implementations obtained from the author and plug them in the GC-RANSAC~\cite{GCRansac2018} framework. The results of the best performing 2PT+D \& 5pt combination are reported in~\cref{tab:results-calib-scannet} and~\cref{tab:megadepth-calib}. The full comparison results (with calibrated cameras, SP+LG matches) are shown in~\cref{tab:rebuttal-scannet}
and~\cref{tab:rebuttal-megadepth}. Our method consistently outperforms the scale-only methods from~\cite{barath2022relative}. In addition, as we mentioned in~\cref{sec:related_work}, we find that the 2PT+D solver suffers from degeneration of using only 2 correspondences due to rank deficiency. \\


\begin{table}[t]
\scriptsize
\setlength{\tabcolsep}{3pt}
\centering
\begin{tabular}{ccccc|ccc} 

\toprule
\multicolumn{2}{c}{\multirow{2}{*}{Method}} & \multirow{2}{*}{MD Model} & \multicolumn{2}{c}{Med. Err. $\downarrow$} & \multicolumn{3}{c}{Pose Error AUC (\%) $\uparrow$} \\
& & & $\varepsilon_{\bm{R}}(\degree)$ & $\varepsilon_{\bm{t}}$(\degree) & @5\degree & @10\degree & @20\degree \\
\midrule
\multirow{6}{*}{From [4]} & 2PT+D \& 4PT+D & DA-met. & 5.31 & 17.65 & 6.42 & 16.32 & 29.90 \\
& 2PT+D \& 4PT+D & MoGe & 4.10 & 14.43 & 8.53 & 20.03 & 34.47 \\
& 2PT+D \& 5pt & DA-met. & 1.90 & 5.67 & 20.62 & 38.45 & 54.94 \\
& 2PT+D \& 5pt & MoGe & 1.88 & 5.66 & 20.74 & 38.53 & 54.98 \\
& Sim. P3P \& 5pt & DA-met. & 1.87 & 5.64 & 20.92 & 38.48 & 54.63 \\
& Sim. P3P \& 5pt & MoGe & 1.90 & 5.76 & 20.59 & 38.20 & 54.43 \\
\midrule
\multirow{2}{*}{Ours} & Ours-calib & DA-met. & \underline{1.68} & \underline{4.97} & \underline{22.41} & \underline{42.18} & \underline{59.96} \\
& Ours-calib & MoGe & \textbf{1.57} & \textbf{4.77} & \textbf{23.36} & \textbf{43.39} & \textbf{61.08} \\
\bottomrule
\end{tabular}
\caption{Comparison with scale-only solvers from~\cite{barath2022relative} with calibrated cameras on ScanNet-1500.}
\label{tab:rebuttal-scannet}
\end{table}

\begin{table}[t]
\scriptsize
\centering
\setlength{\tabcolsep}{3pt}
\begin{tabular}{ccccc|ccc}

\toprule
\multicolumn{2}{c}{\multirow{2}{*}{Method}} & \multirow{2}{*}{MD Model} & \multicolumn{2}{c}{Med. Err. $\downarrow$} & \multicolumn{3}{c}{Pose Error AUC (\%) $\uparrow$} \\
& & & $\varepsilon_{\bm{R}}(\degree)$ & $\varepsilon_{\bm{t}}(\degree)$ & @5\degree & @10\degree & @20\degree \\
\midrule
\multirow{6}{*}{From [4]} & 2PT+D \& 4PT+D & DA-met. & 6.84 & 25.54 & 8.89 & 16.64 & 26.96 \\
& 2PT+D \& 4PT+D & MoGe & 3.54 & 14.56 & 13.72 & 23.74 & 36.20 \\
& 2PT+D \& 5pt & DA-met. & 0.52 & 1.37 & 57.83 & 72.85 & 83.73 \\
& 2PT+D \& 5pt & MoGe & 0.54 & 1.44 & 55.86 & 70.97 & 82.20 \\
& Sim.P3P \& 5pt & DA-met. & 0.49 & 1.34 & 58.23 & 72.97 & 83.68 \\
& Sim.P3P \& 5pt & MoGe & 0.57 & 1.52 & 55.99 & 71.14 & 82.26 \\
\midrule
\multirow{2}{*}{Ours}
& Ours-calib & DA-met. & \underline{0.47} & \underline{1.26} & \underline{59.80} & \underline{74.77} & \underline{85.47} \\
& Ours-calib & MoGe & \textbf{0.41} & \textbf{1.16} & \textbf{63.48} & \textbf{77.79} & \textbf{87.18} \\

\bottomrule
\end{tabular}
\caption{Comparison with scale-only solvers from~\cite{barath2022relative} with calibrated cameras on MegaDepth-1500.}
\label{tab:rebuttal-megadepth}
\end{table}

\begin{figure*}[tb]
\scriptsize
\centering
\setlength\tabcolsep{3pt} 

\begin{minipage}{0.95\linewidth}
\renewcommand{\arraystretch}{0.4}
\begin{tabular}{cccccc}
\includegraphics[width=0.13\linewidth]{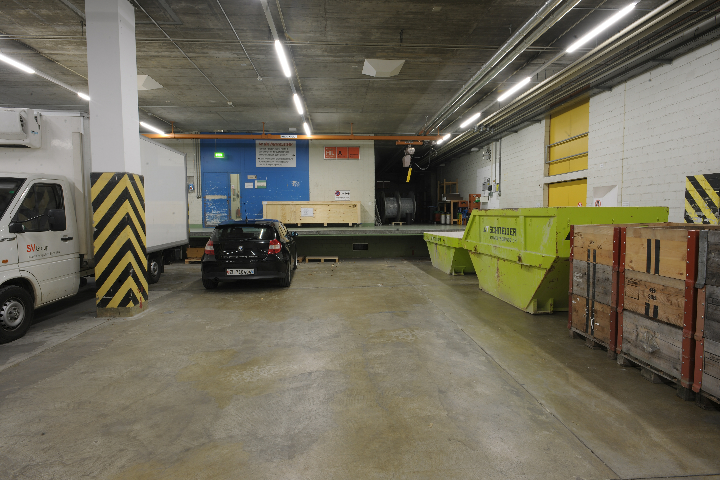} & & &
\raisebox{-0.5\height}[0pt][0pt]{\includegraphics[width=0.27\linewidth]{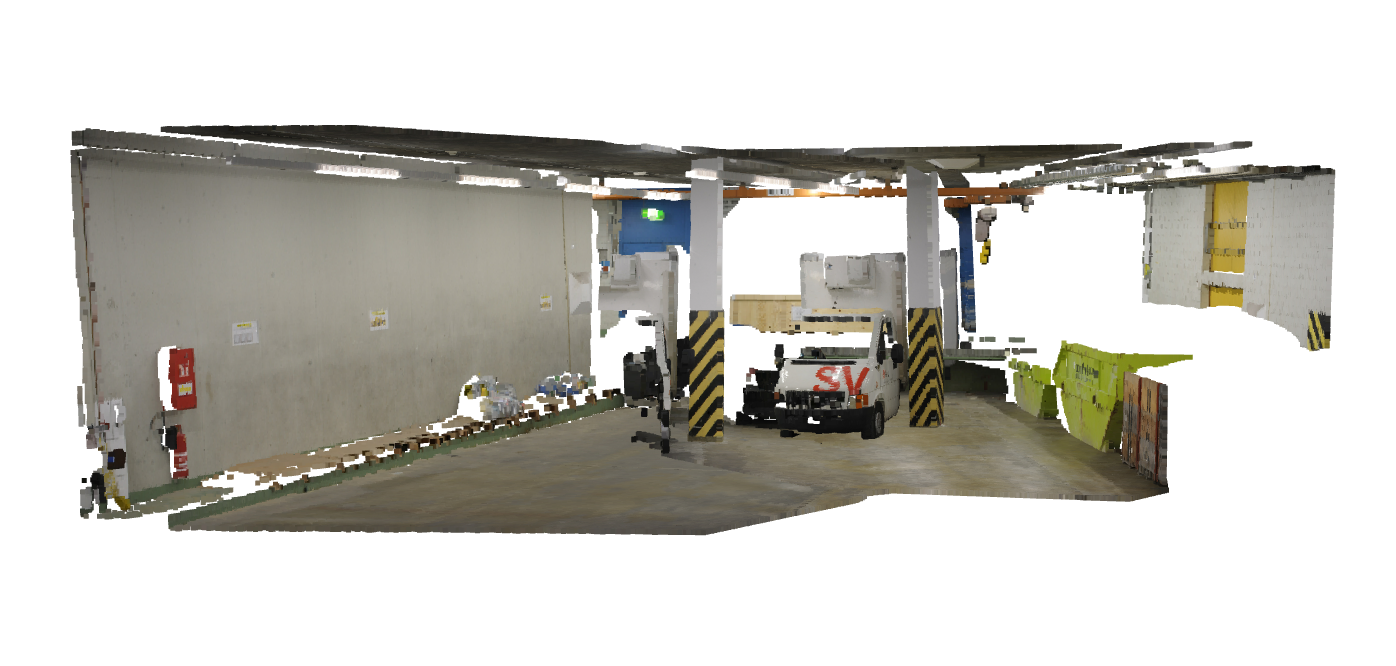}} &
\raisebox{-0.5\height}[0pt][0pt]{\includegraphics[width=0.27\linewidth]{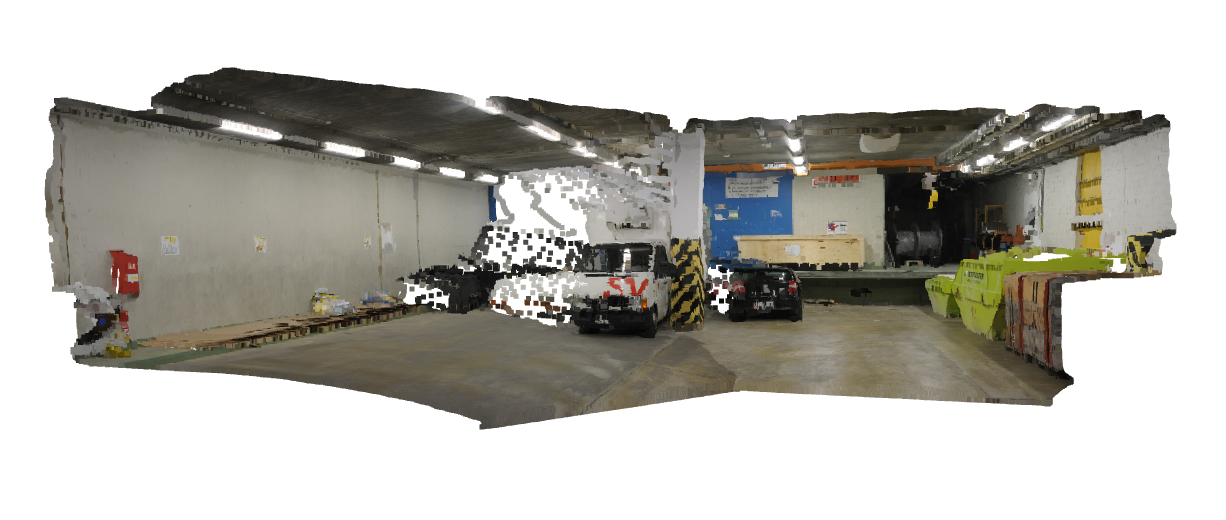}} &
\raisebox{-0.5\height}[0pt][0pt]{\includegraphics[width=0.27\linewidth]{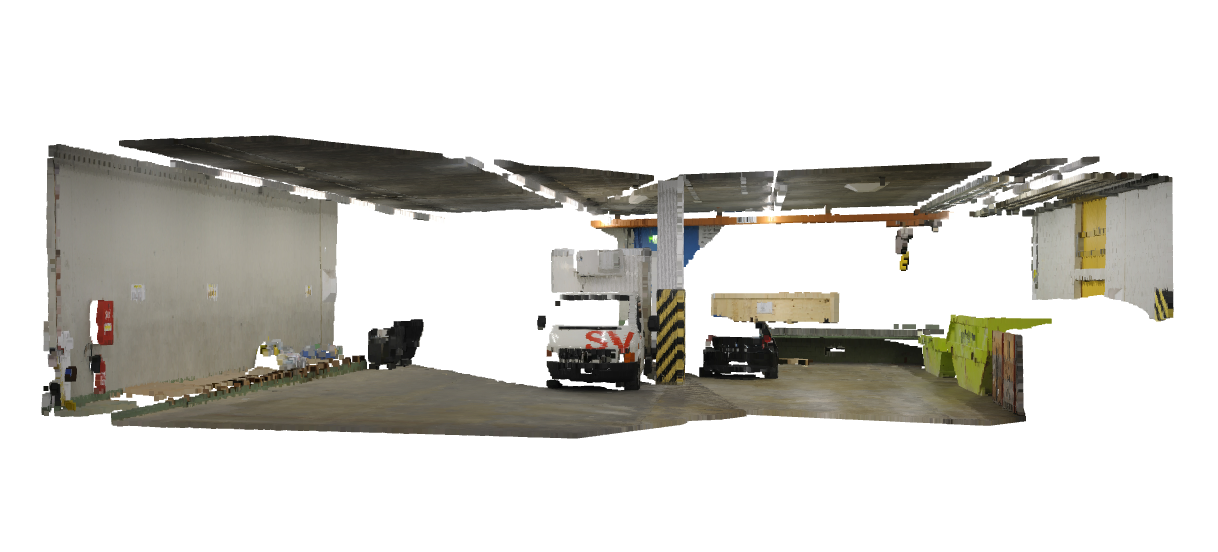}} \\
\includegraphics[width=0.13\linewidth]{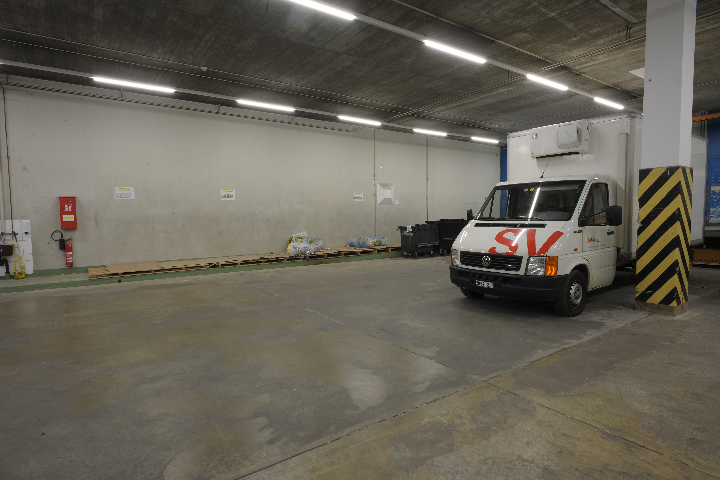} & \\[1em]
\includegraphics[width=0.13\linewidth]{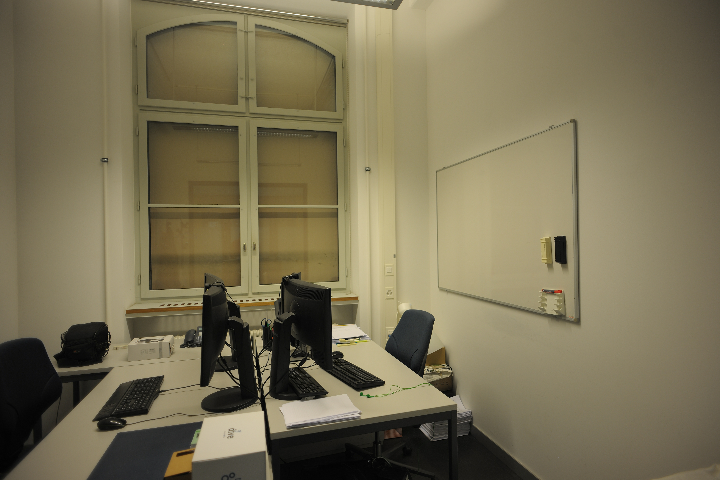} & & &
\raisebox{-0.5\height}[0pt][0pt]{\includegraphics[width=0.27\linewidth]{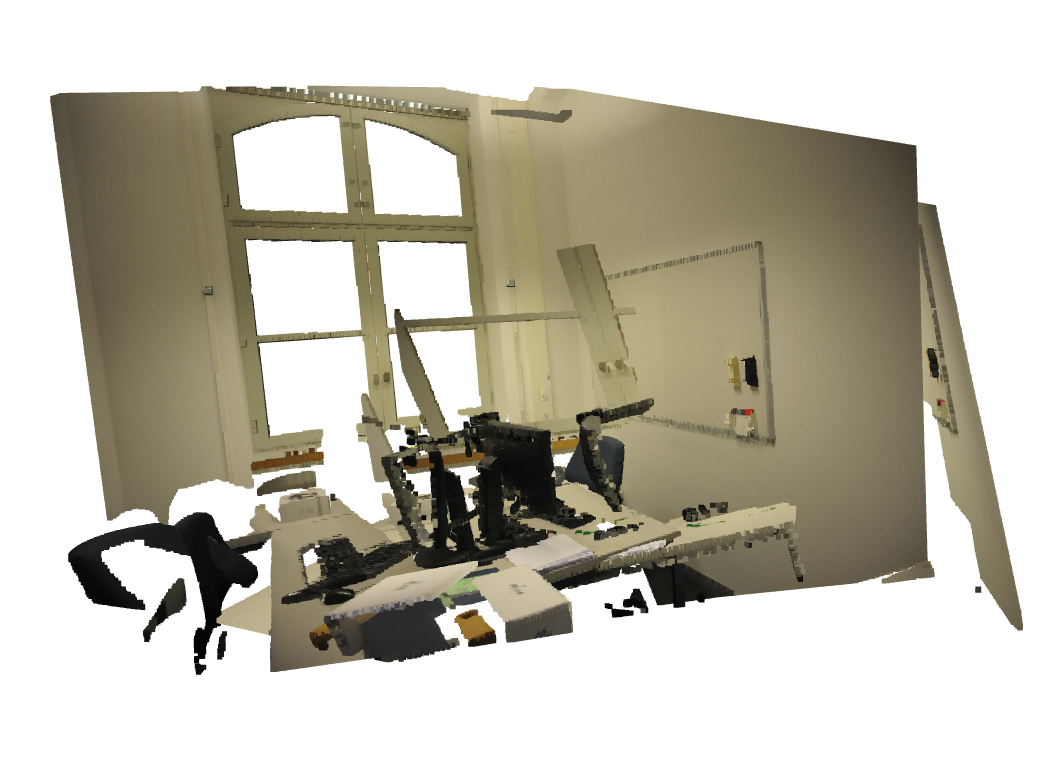}} &
\raisebox{-0.5\height}[0pt][0pt]{\includegraphics[width=0.27\linewidth]{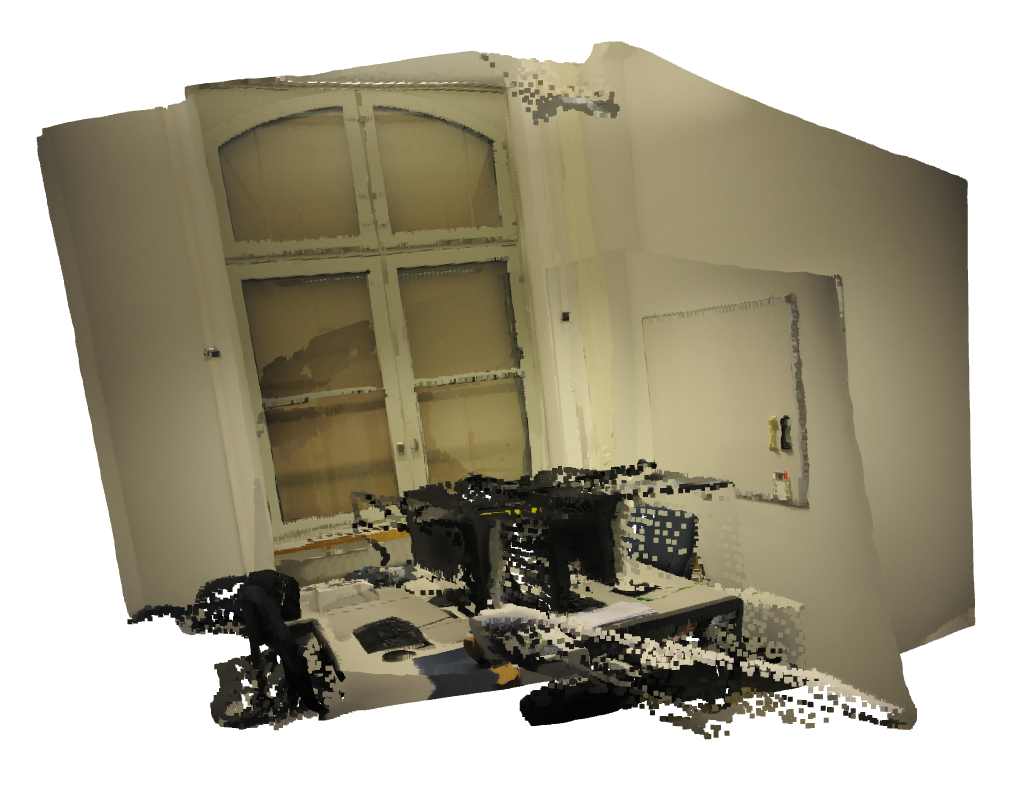}} &
\raisebox{-0.5\height}[0pt][0pt]{\includegraphics[width=0.27\linewidth]{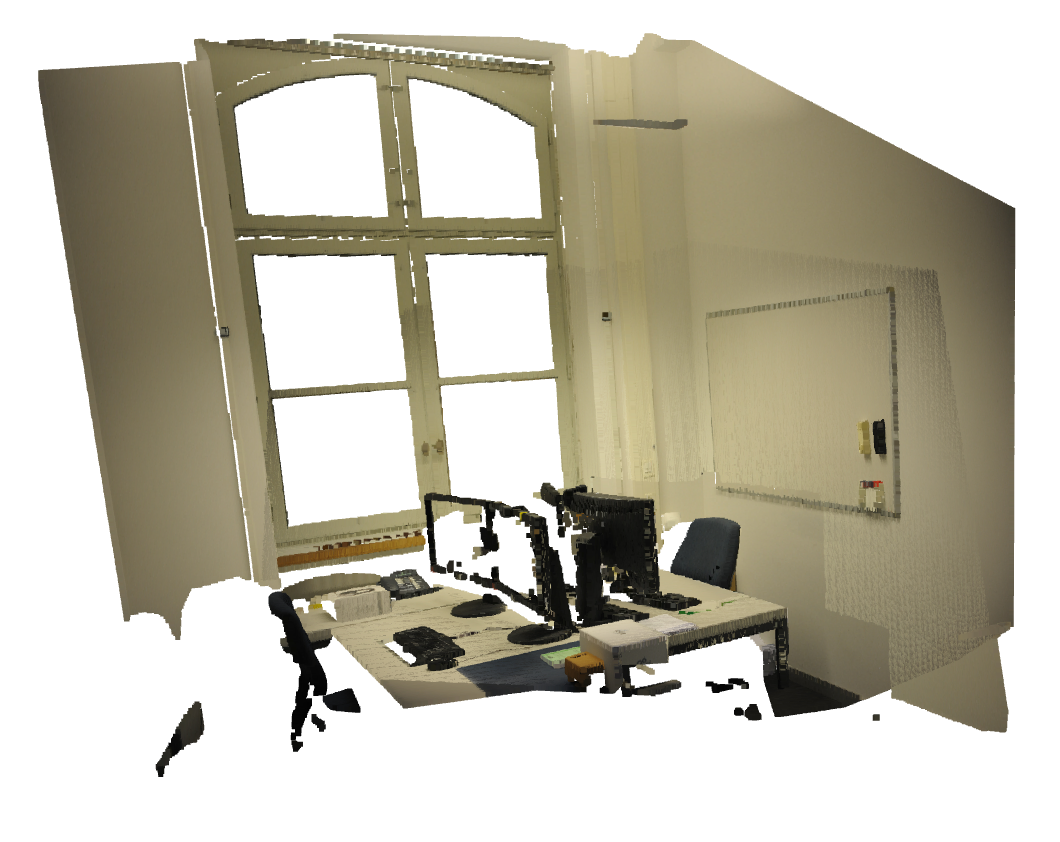}} \\
\includegraphics[width=0.13\linewidth]{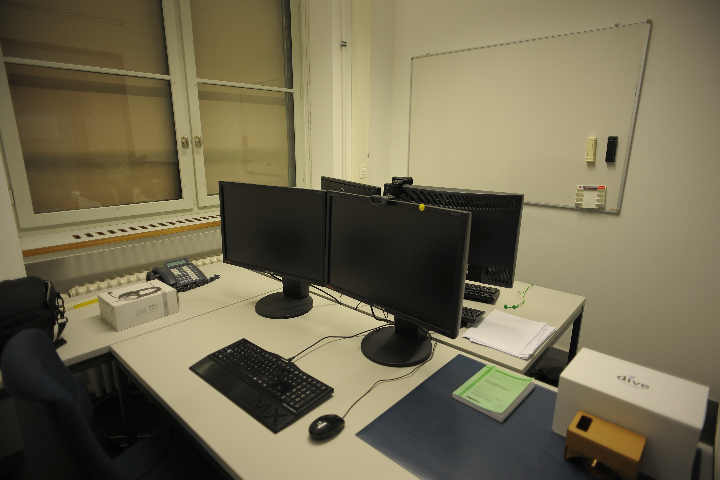} & \\
\end{tabular}
\end{minipage} 
\caption{Additional visualizations on ETH3D~\cite{ETH3D} with shared-focal setting. \textbf{Left:} back-projected GT depth with relative pose found by PoseLib-6pt~\cite{PoseLib} and translation rescaled to match scale with GT translation; \textbf{Middle:} back-projected depth priors from Marigold~\cite{marigold} aligned using the output scale, shifts, relative pose, and focal length from our method; \textbf{Right:} Aligned GT depth with GT pose.}
\label{fig:vis_supp_eth3d}
\end{figure*}

\begin{figure*}[tb]
\centering
\setlength\tabcolsep{3pt} 

\renewcommand{\arraystretch}{0.4}
\begin{tabular}{ccccc}
\includegraphics[width=0.13\linewidth]{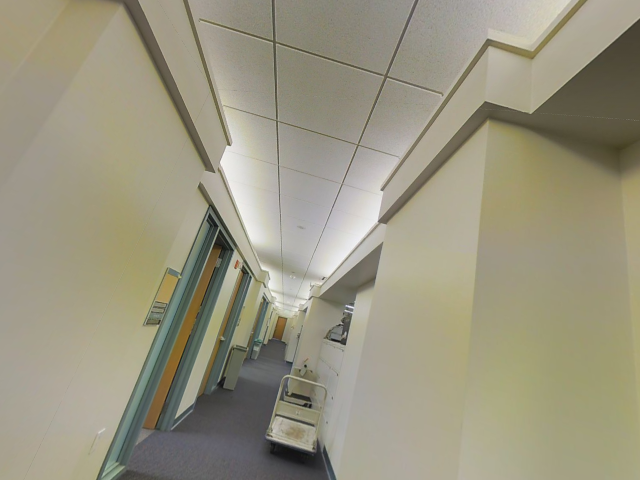} & & &
\raisebox{-0.5\height}[0pt][0pt]{\includegraphics[width=0.27\linewidth]{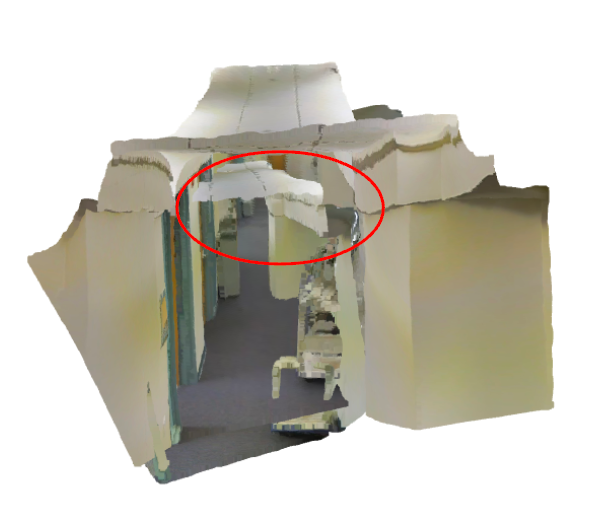}} &
\raisebox{-0.5\height}[0pt][0pt]{\includegraphics[width=0.27\linewidth]{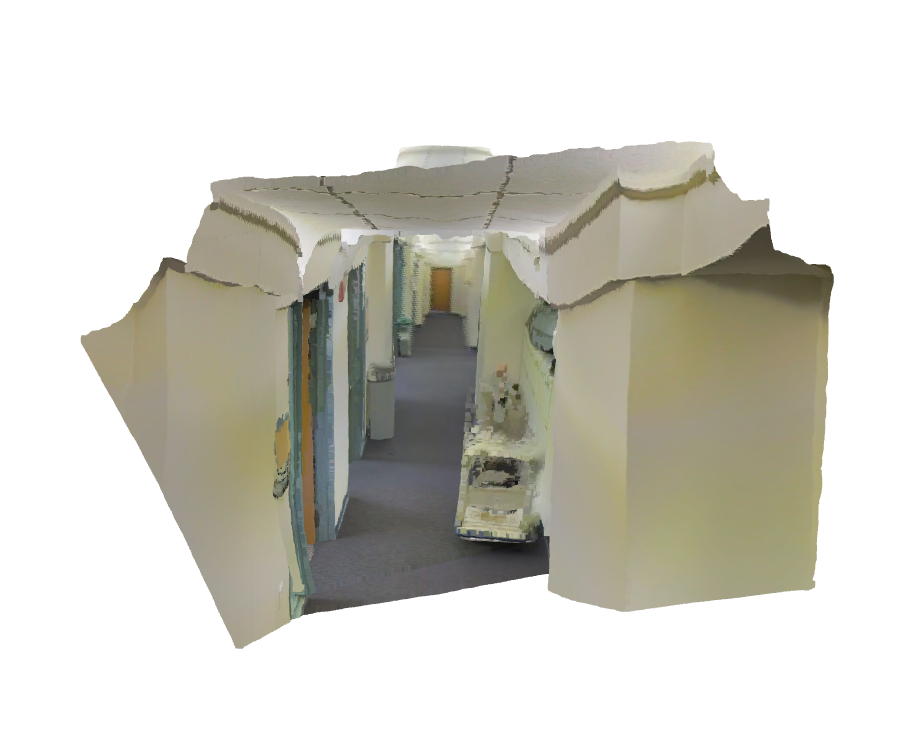}} \\
\includegraphics[width=0.13\linewidth]{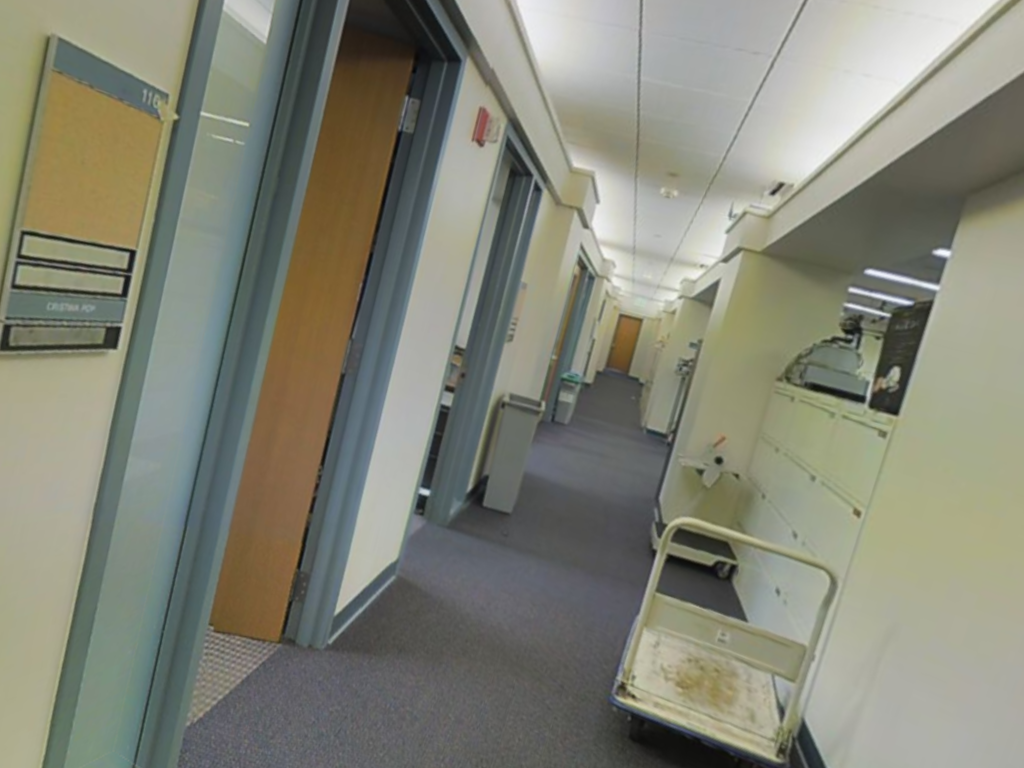}  \\
\includegraphics[width=0.13\linewidth]{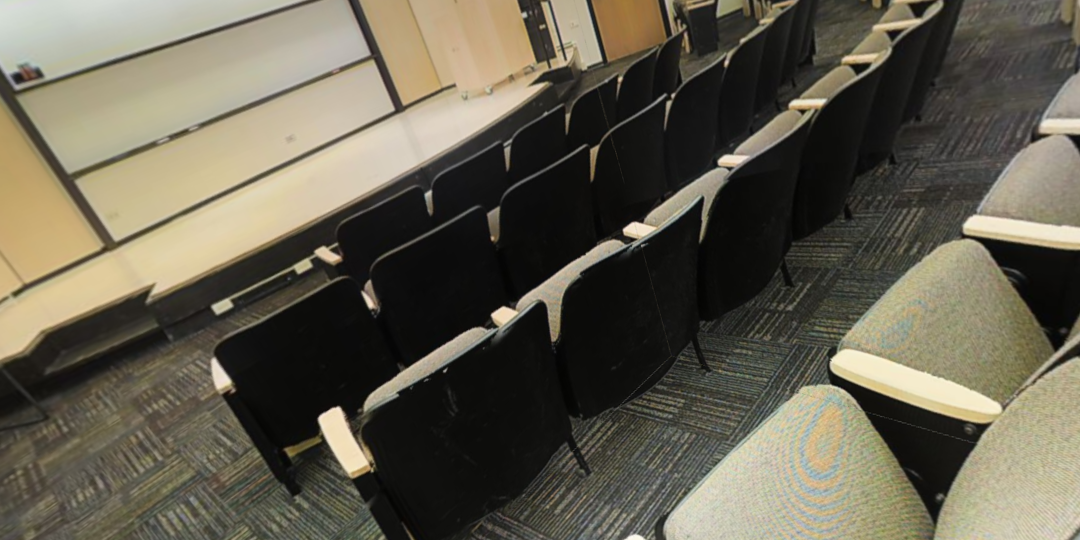} & & &
\raisebox{-0.5\height}[0pt][0pt]{\includegraphics[width=0.27\linewidth]{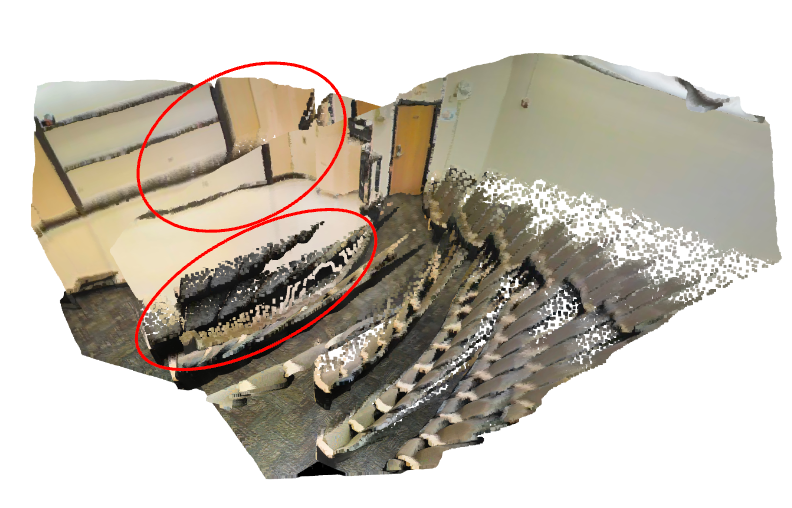}} &
\raisebox{-0.5\height}[0pt][0pt]{\includegraphics[width=0.27\linewidth]{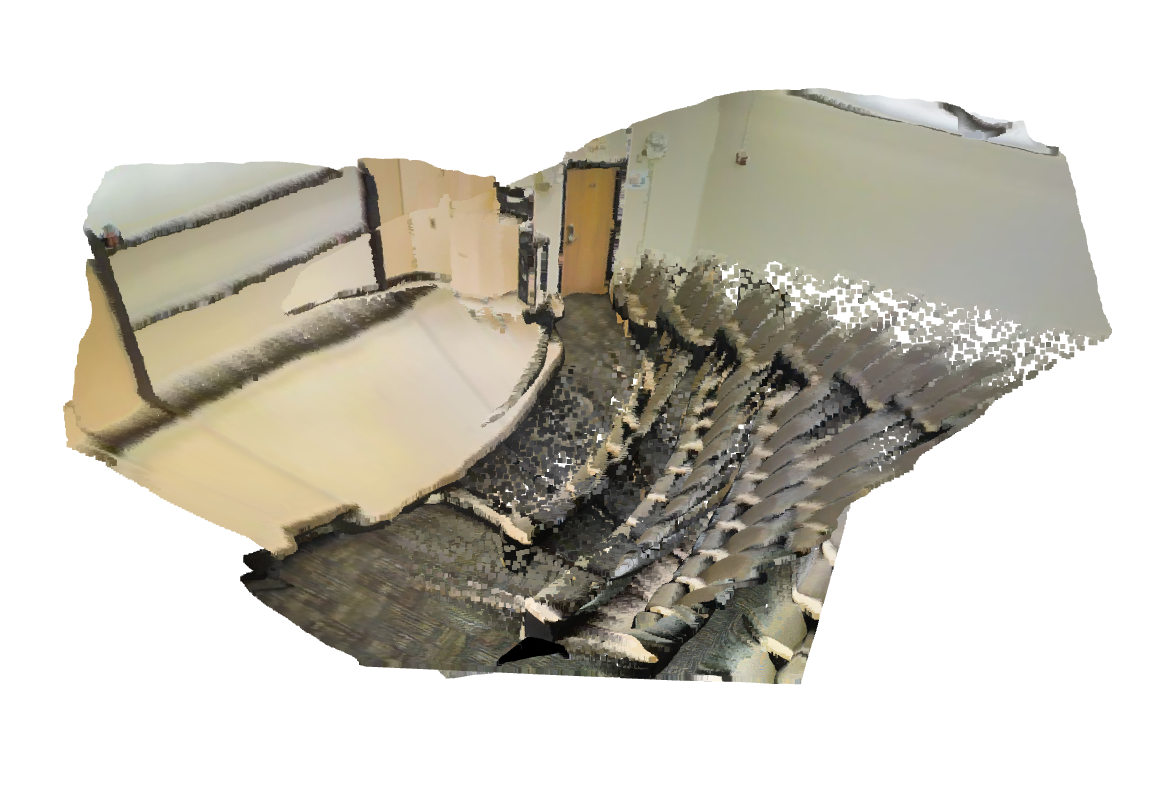}} \\
\includegraphics[width=0.13\linewidth]{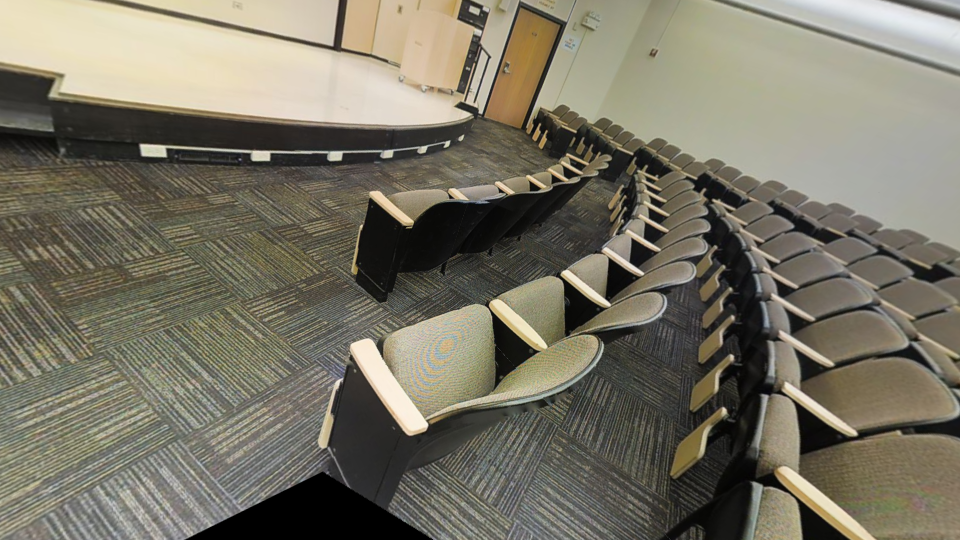} & \\
\end{tabular}

\caption{Additional visualizations on Stanford 2D-3D-S~\cite{2d3ds} image pairs with two-focal setting. \textbf{Left:} back-projected depth priors aligned using the relative pose found by PoseLib-7pt~\cite{PoseLib} baseline; \textbf{Right:} back-projected depth priors aligned using the relative pose found by our two-focal estimator. Both point clouds are aligned using the scale and shifts from our method, but focal lengths from each method, with translation found by the point-based baseline is rescaled to match the length of translation found by our method. (Currently no GT depth available for the sampled image pairs.)}
\label{fig:vis_supp_2d3ds}
\end{figure*}

\begin{figure*}[tb]
\scriptsize
\centering
\setlength\tabcolsep{3pt} 

\begin{minipage}{.95\linewidth}
\renewcommand{\arraystretch}{0.4}
\begin{tabular}{cccccc}
\includegraphics[width=0.13\linewidth]{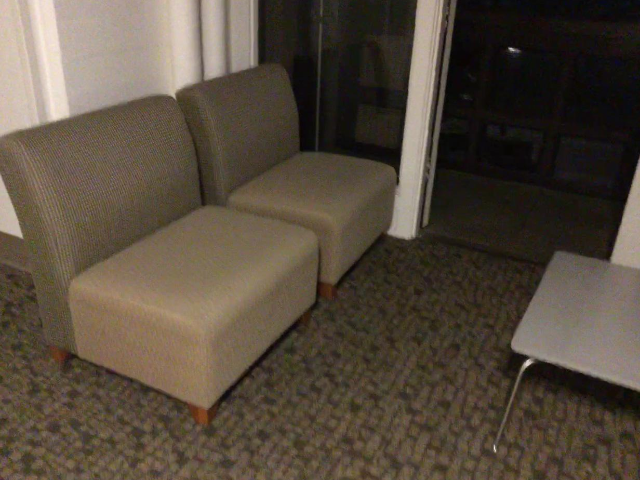} & & &
\raisebox{-0.5\height}[0pt][0pt]{\includegraphics[width=0.27\linewidth]{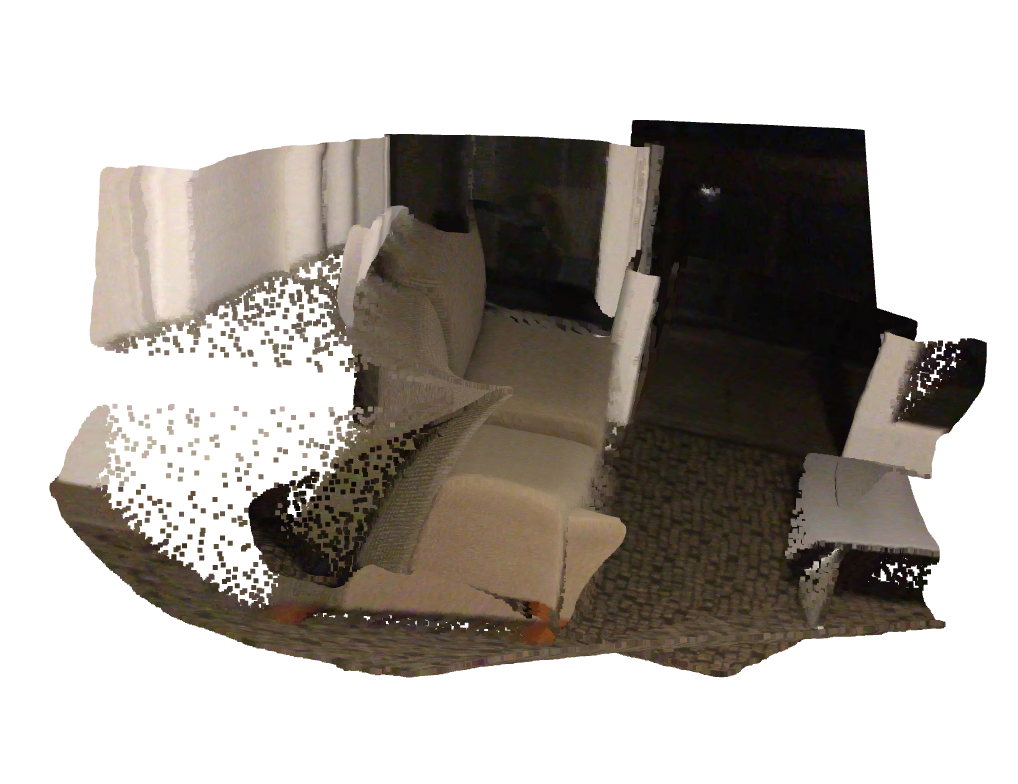}} &
\raisebox{-0.5\height}[0pt][0pt]{\includegraphics[width=0.27\linewidth]{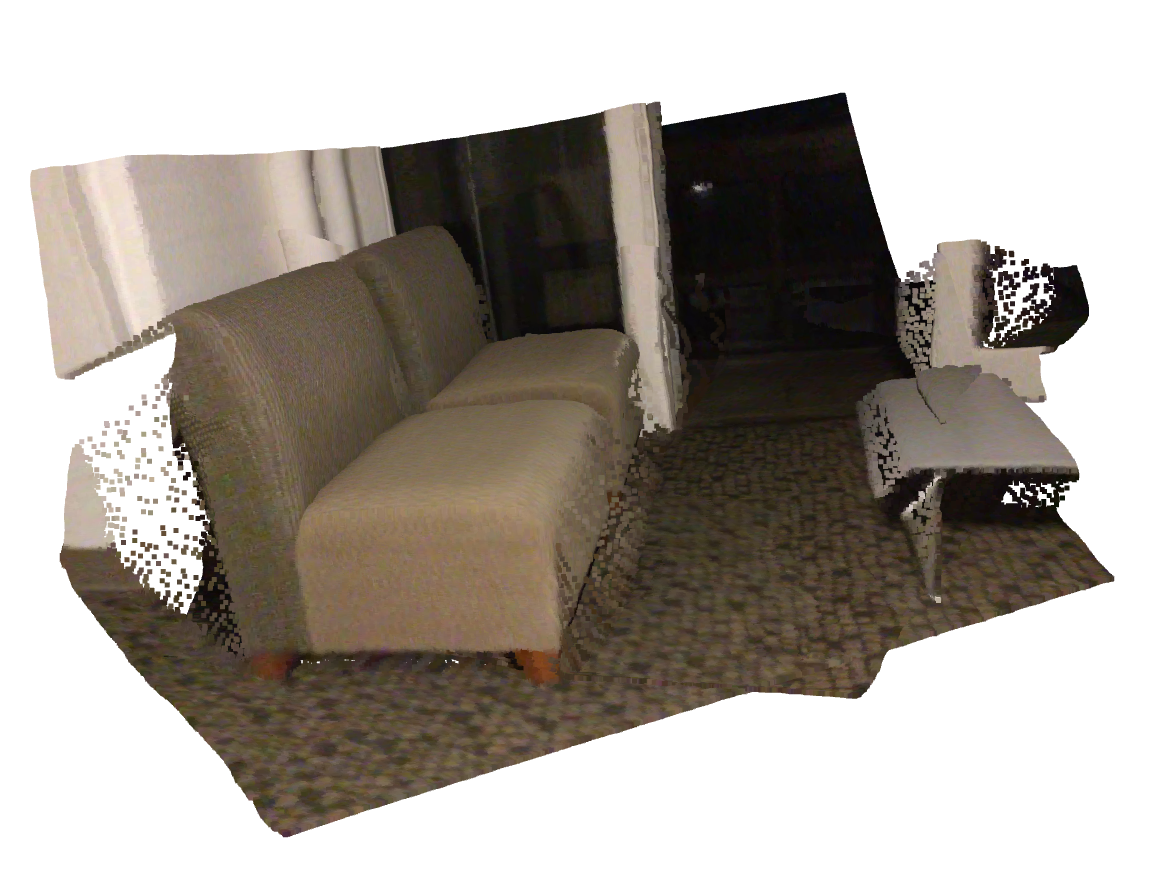}} &
\raisebox{-0.5\height}[0pt][0pt]{\includegraphics[width=0.27\linewidth]{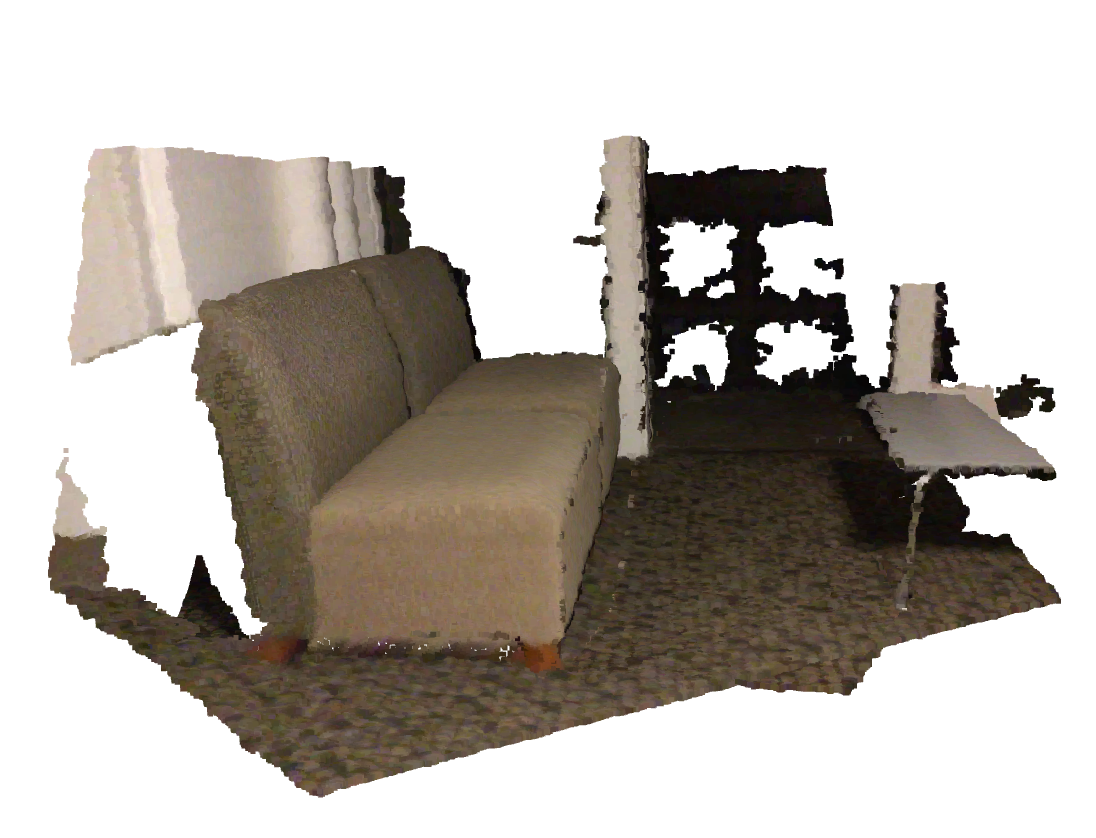}} \\
\includegraphics[width=0.13\linewidth]{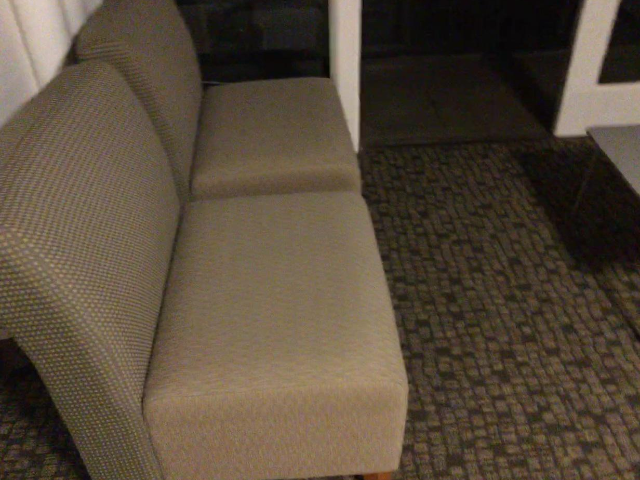} \\[1em]
\includegraphics[width=0.13\linewidth]{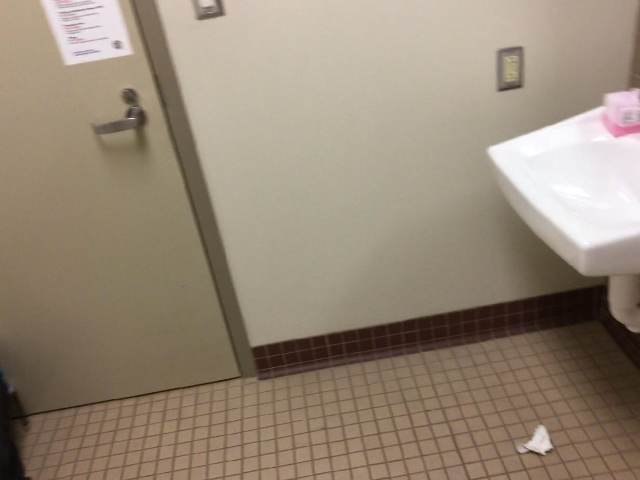} & & &
\raisebox{-0.5\height}[0pt][0pt]{\includegraphics[width=0.27\linewidth]{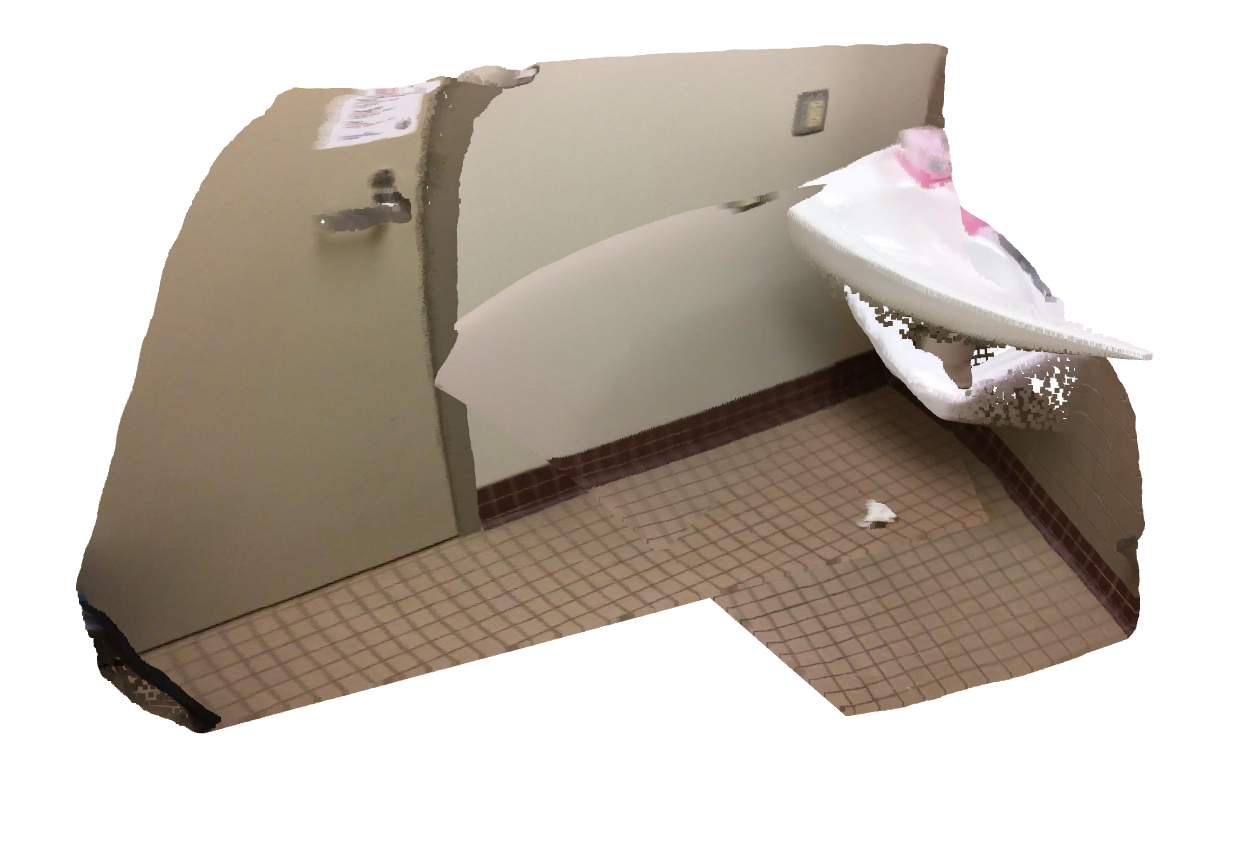}} &
\raisebox{-0.5\height}[0pt][0pt]{\includegraphics[width=0.27\linewidth]{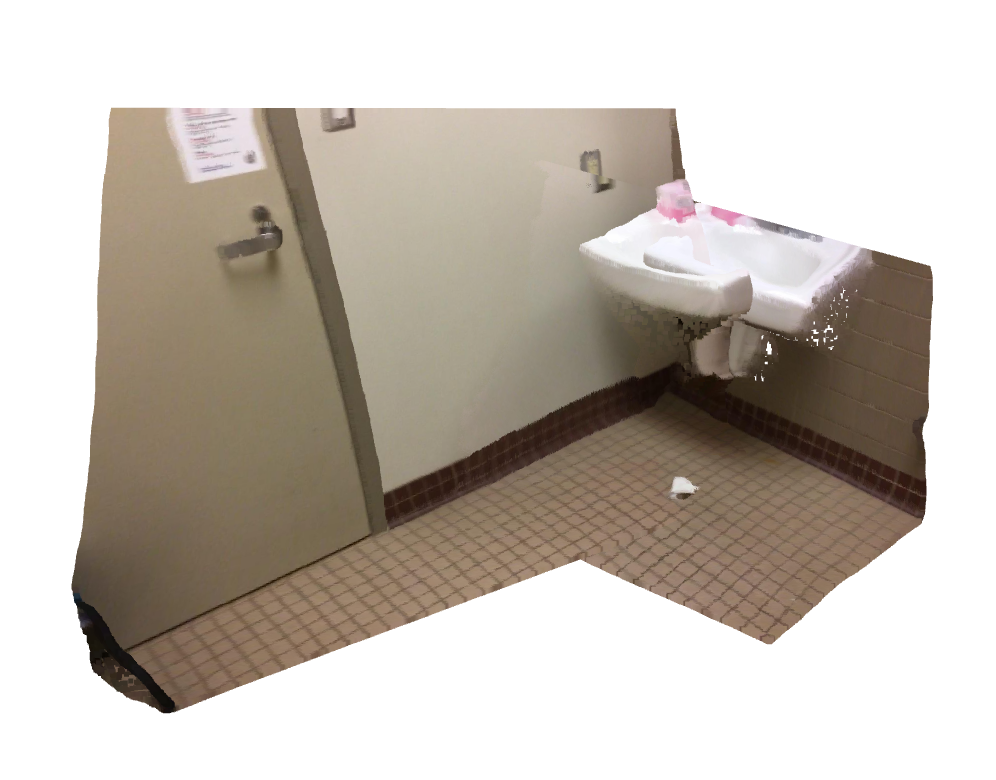}} &
\raisebox{-0.5\height}[0pt][0pt]{\includegraphics[width=0.27\linewidth]{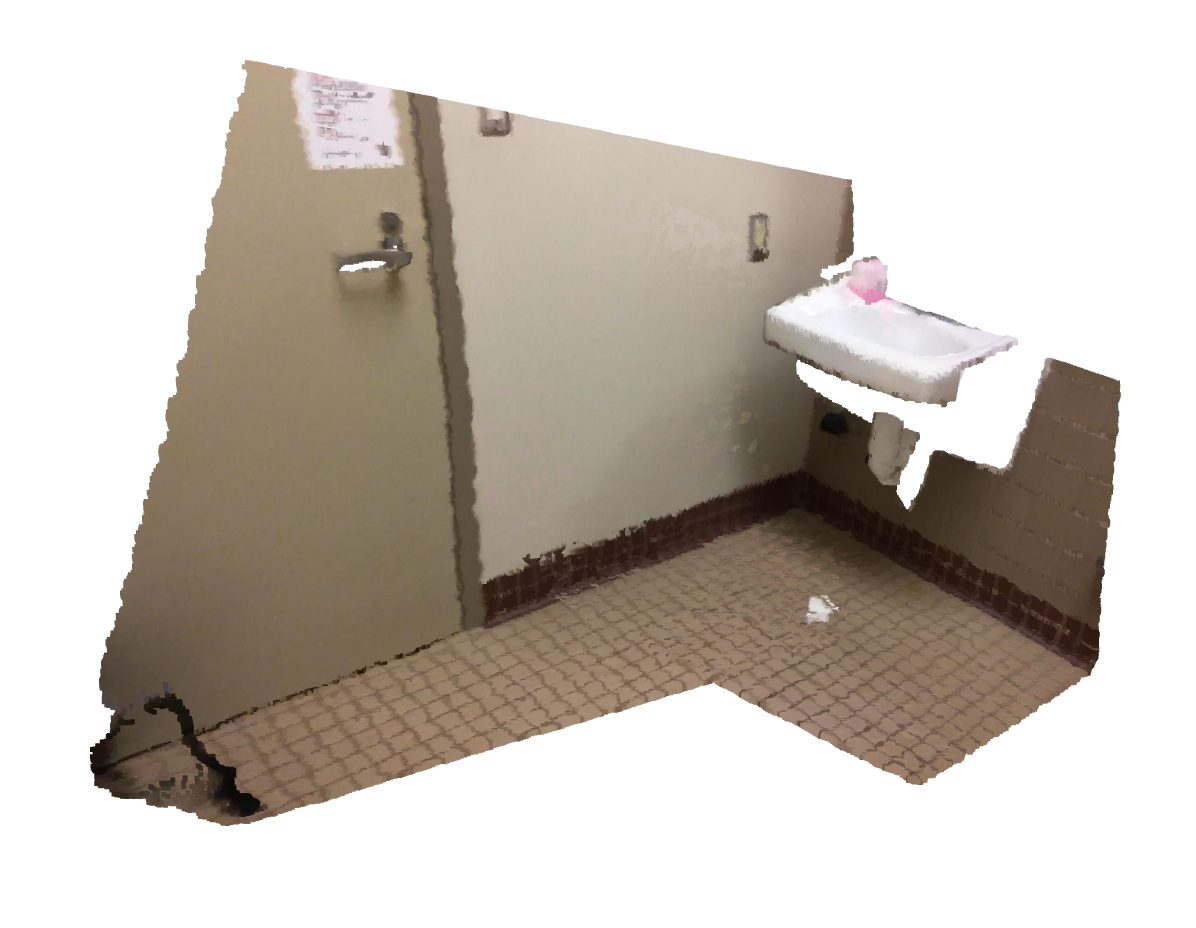}} \\
\includegraphics[width=0.13\linewidth]{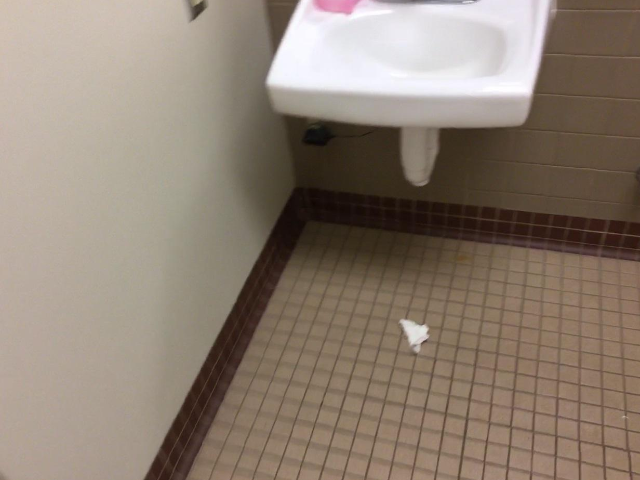} \\[1em]
\includegraphics[width=0.13\linewidth]{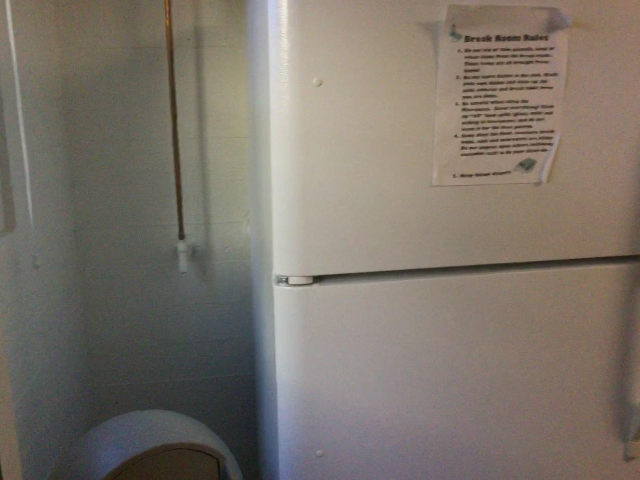} & & &
\raisebox{-0.5\height}[0pt][0pt]{\includegraphics[width=0.27\linewidth]{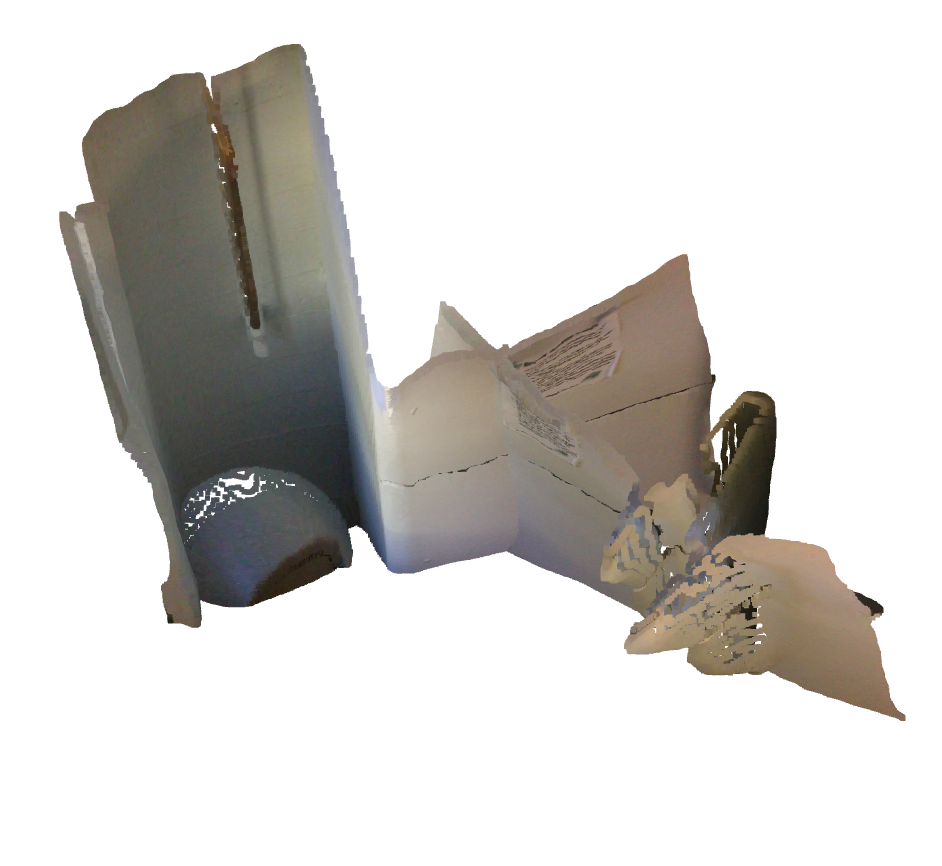}} &
\raisebox{-0.5\height}[0pt][0pt]{\includegraphics[width=0.27\linewidth]{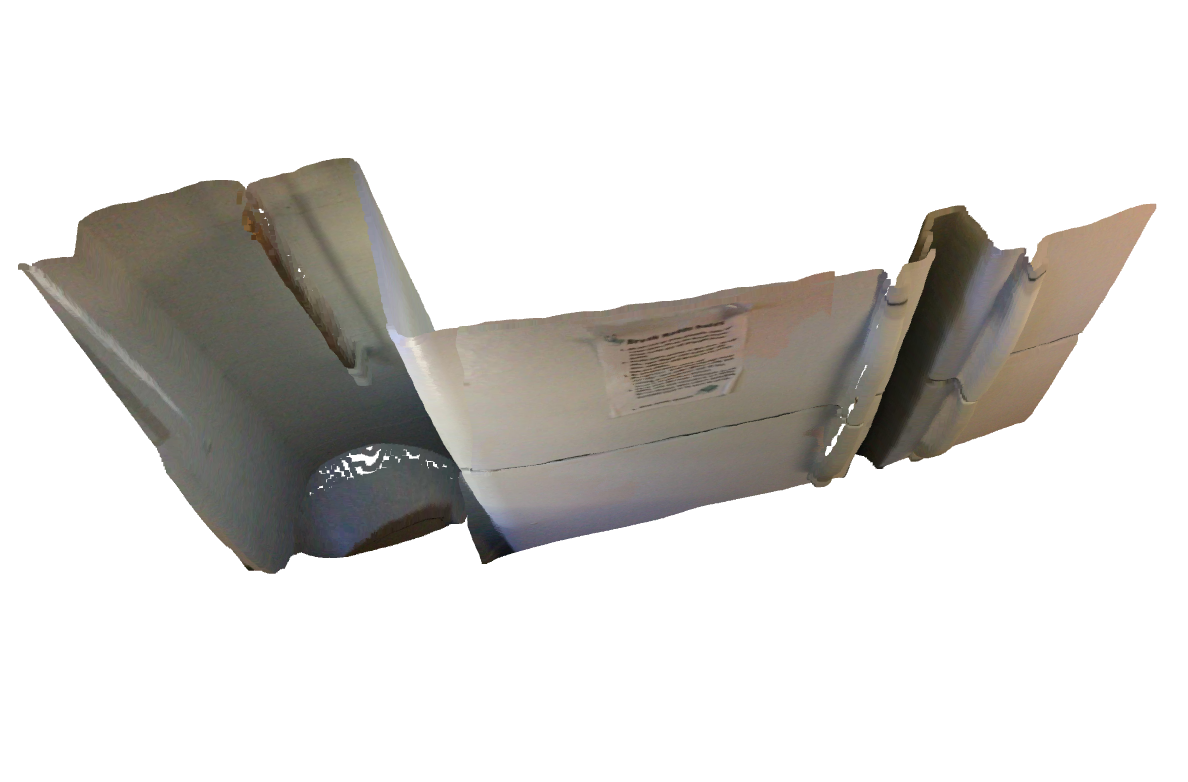}} &
\raisebox{-0.5\height}[0pt][0pt]{\includegraphics[width=0.27\linewidth]{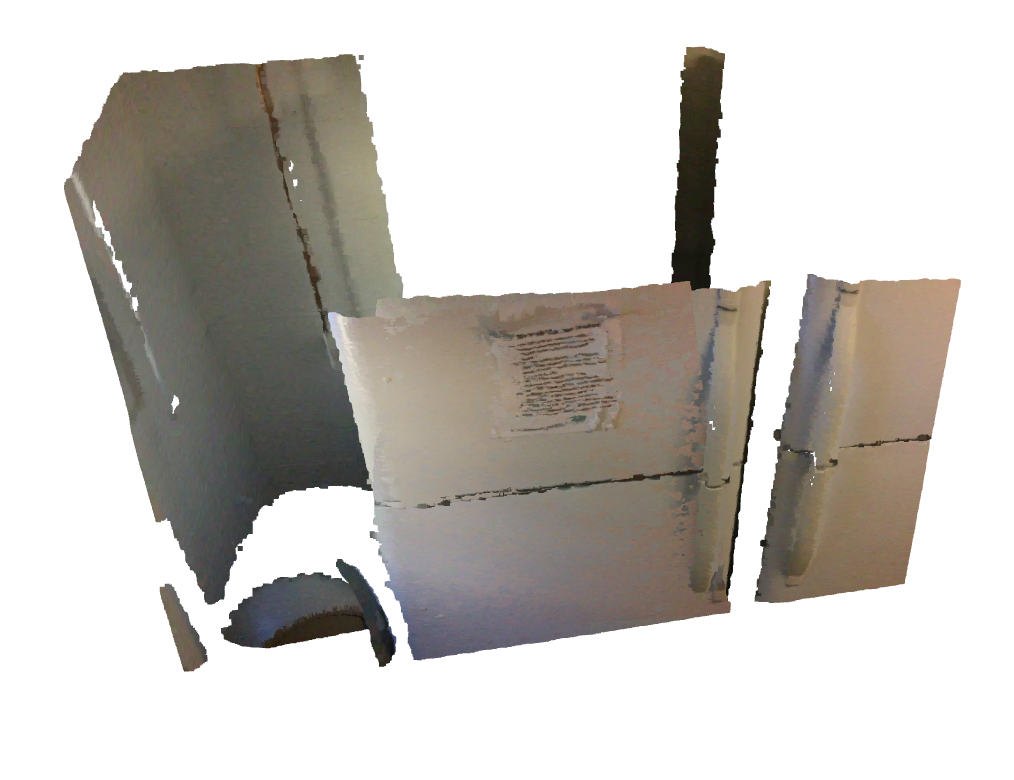}} \\
\includegraphics[width=0.13\linewidth]{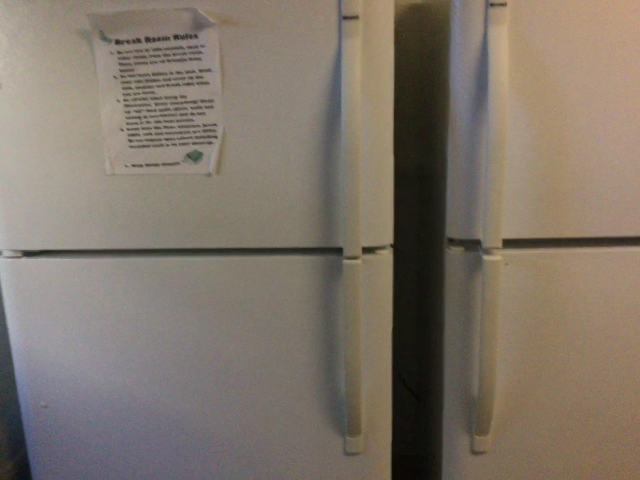} \\
\end{tabular}
\end{minipage} 
\caption{Visualization of aligned point clouds on image pairs from ScanNet-1500\cite{dai2017scannet} using: \textbf{Left:} the scale-only ablated baseline (\cref{sec:model_scale_and_shift}); \textbf{Middle:} our method (calibrated setting); \textbf{Right:} GT depth.}
\label{fig:vis_supp_scale_only}
\end{figure*}

\noindent\textbf{Additional visual results: }
We provide more visualization examples in addition to~\cref{fig:vis_samples}. In~\cref{fig:vis_supp_eth3d} and~\cref{fig:vis_supp_2d3ds} we show examples on ETH3D~\cite{ETH3D} with the shared-focal setting, and on 2D-3D-S~\cite{2d3ds} images with the two-focal setting. By incorporating monocular depth priors, our method is able to find more accurate pose together with scale and shifts of the depth priors that lead to better and more correct alignment of the back-projected point clouds. In~\cref{fig:vis_supp_scale_only} we show examples on ScanNet~\cite{dai2017scannet} comparing to the scale-only ablated baseline as described in~\cref{sec:model_scale_and_shift}. Only modeling scale without the shift can lead to failure cases with incorrect alignment and distortion visible in the aligned point clouds.

\section{Additional Discussion on Proposed Solvers}
In~\cref{sec:solvers} we mentioned the proposed calibrated solver is minimal with 3 point correspondences and related depth priors while the shared-focal and two-focal solvers are non-minimal. We provide in this section a simple reasoning of the minimality of the calibrated solver, and discuss about possible minimal versions for the shared-focal and two-focal solvers. \\

\noindent\textbf{Minimality of Calibrated Solver:} The calibrated solver takes 3 point correspondences and depth priors to solve for the relative pose $\bm{R}, \bm{t}$, depth scale $\alpha$ and shifts $\beta_1, \beta_2$. Conventionally, relative pose are solved by finding the essential matrix which has 5 degrees-of-freedom (DOFs) up to an unknown scale. In our setup, however, because the solved relative pose (translation) has a fixed scale consistent with the solved depth scale and shifts, the relative pose now has 6 DOFs. In total the problem has $6 + 1 + 2 = 9$ DOFs, and is minimally solvable with 3 point correspondences and depth priors since each pair of 2D correspondences gives 1 epipolar constraint, and with depth priors we can additionally have 2 projection constraints per pair. \\

\noindent\textbf{Shared-focal and Two-focal Solvers:}
For the two solvers that we propose for uncalibrated cases, the problem is not minimal and our solvers ignore 1 or 2 of the 6 constraints we have.
This means that the solutions we get might not exactly satisfy the correspondences in the sample set (which can be later taken care by the hybrid RANSAC pipeline).
Another approach would be to drop some of the input data, instead of dropping equations.
For example, one could take 3 pairs of point matches with depth and one pair without depth, or with partial depth (only in one view). 
One approach to formulate this would then be to parameterize the \textit{missing depth} as extra unknowns.
We briefly explored this option but applying \cite{larsson2017efficient} yielded solvers with elimination templates of size $360\times 374$ (14 solutions) and $716 \times 744$ (28 solutions), which are too slow to be used in practice.

\section{Limitation and Future Work}
We discuss here some limitations of our affine modeling of the monocular depth priors and the proposed pipeline, improvements of which could lead to interesting and promising future works.

First, while our affine correction of the monocular depth priors is proven beneficial for estimating relative pose and outperforms previous methods that only model the scale, the affine modeling of depth maps is simple and limited with only two parameters (scale and shift). In practice, we have found that the estimated $\beta$ might not uniformally agree with all pixels and their depth priors, but rather different groups of regions/surfaces in the image can be better fitted with different shift values. This is due to the fact that MDE models are better at inferring relative depth between pixels of the same object/surface than pixels across different surfaces due to the ambiguous scales among objects. We also observed that the same depth map could result in a few different groups of $\beta$ values when estimating relative pose with different images (all with good estimated poses), depending on the different groups of regions/surfaces that are aligned by the inlier correspondences. Therefore, one interesting future work direction would be to enhance the affine modeling to more fine-grained region-based modeling, possibly with the help of the latest advances in image segmentation. At the same time, our method can also benefit from more advanced monocular depth models with better accuracy on outdoor images or inter-image consistencies as can be seen from~\cref{tab:ablat-monodepth}.

Second, while we are able to get good results by empirically setting $\lambda_s$ to 1.0 in the experiments, a more mathematically sound way of balancing between the depth-induced reprojection errors and Sampson error could be developed. This can be especially beneficial when the depth priors are less reliable (\textit{e.g.} on outdoor images), and could utilize information such as uncertainty modeling of the depth priors and inlier ratios of the different types of correspondences.

Third, our proposed pipeline is dependent on pixel correspondences produced by off-the-shelf matchers, and therefore only limited part of the estimated depth priors are utilized. It would be interesting to explore whether depth priors of other unmatched pixels could provide additional geometric constraints. 

Lastly, a natural extension of our pipeline is to extend our affine modeling to multi-view, possibly through bundle-adjustment to solve multi-view problems like structure-from-motion.

%% file: tables/ablation_monodepth.tex
\begin{table}[tb]
\scriptsize
\centering
\setlength{\tabcolsep}{3pt}

\begin{tabular}{ccccc|ccc}
\toprule
\multirow{2}{*}{Task} & \multirow{2}{*}{\makecell{Method / \\ MD Model}} & \multicolumn{3}{c}{Med. Err. $\downarrow$} & \multicolumn{3}{c}{Pose Err. AUC (\%) $\uparrow$} \\
& & $\varepsilon_{\bm{R}}(\degree)$ & $\varepsilon_{\bm{t}}(\degree)$ & $\varepsilon_f$(\%) & @5\degree & @10\degree & @20\degree \\

\midrule
\multirow{8}{*}{\makecell{ScanNet-1500 \\ Calibrated}} & PoseLib-5pt & 2.08 & 5.44 & - & 19.48 & 38.09 & 56.08 \\ 
\cmidrule{2-8}
& Omnidata & 1.76 & 5.42 & - & 21.24 & 39.74 & 57.75 \\
& Marigold & 1.72 & 5.28 & - & 21.18 & 40.38 & 58.13 \\
& DA-met. & \underline{1.68} & \underline{4.97} & - & \underline{22.41} & \underline{42.18} & \underline{59.96} \\
& DAv2 (inv) & 1.90 & 5.72 & - & 19.60 & 38.10 & 56.18 \\
& DAv2-met. & 1.72 & 5.25 & - & 21.90 & 41.01 & 59.16 \\
& MoGe & \textbf{1.57} & \textbf{4.77} & - & \textbf{23.36} & \textbf{43.39} & \textbf{61.08} \\

\cmidrule{2-8}
& GT Depth & 1.54 & 4.57 & - & 25.03 & 45.06 & 61.85 \\

\midrule
\multirow{7}{*}{\makecell{ETH3D \\ Shared-focal}} & PoseLib-6pt & 0.90 & 1.81 & 8.79 & 46.29 & 56.79 & 65.43 \\
\cmidrule{2-8}
& Omnidata & 1.09 & 2.42 & 9.78 & 41.90 & 54.75 & 65.38 \\
& Marigold & \underline{0.94} & \underline{1.79} & \underline{8.06} & \underline{45.55} & \underline{58.91} & \underline{68.71} \\
& DA-met. & 1.06 & 2.37 & 10.47 & 41.81 & 53.33 & 62.94 \\
& DAv2 (inv) & 1.26 & 2.84 & 9.85 & 38.50 & 52.74 & 64.85 \\
& DAv2-met. & \textbf{0.86} & \textbf{1.78} & \textbf{7.71} & \textbf{47.60} & \textbf{59.60} & \textbf{69.41} \\
\cmidrule{2-8}
& GT Depth & 0.36 & 0.81 & 2.07 & 62.26 & 70.52 & 75.90 \\

\midrule
\multirow{7}{*}{\makecell{MegaDepth-1500 \\ Two-focal}} & PoseLib-7pt & 1.97 & 5.72 & 23.64 & 21.23 & 36.80 & 54.89  \\
\cmidrule{2-8}
& Omnidata & \underline{1.53} & \underline{5.41} & \underline{18.60} & \underline{22.67} & \underline{39.94} & \underline{59.15} \\
& Marigold & 2.03 & 6.72 & 23.61 & 18.56 & 34.01 & 53.46 \\
& DA-met. & \textbf{1.25} & \textbf{4.81} & \textbf{15.99} & \textbf{25.70} & \textbf{42.90} & \textbf{61.79} \\
& DAv2 (inv) & 1.62 & 5.69 & 18.93 & 21.50 & 38.46 & 57.04 \\
& DAv2-met. & 2.06 & 7.45 & 24.53 & 18.05 & 32.44 & 50.33 \\
\cmidrule{2-8}
& GT Depth & 0.48 & 3.32 & 6.43 & 38.04 & 54.85 & 70.08 \\

\bottomrule
\end{tabular}
\caption{Results with different MDE models on three tasks on three different datasets. All results are with SP+LG matches. Best results among the different models on each task are \textbf{bolded}, and second best \underline{underlined}.}
\label{tab:ablat-monodepth}
\end{table}